\documentclass[10pt,twocolumn,letterpaper]{article}

\usepackage{booktabs}
\usepackage{multirow}
\usepackage{cvpr}
\usepackage{times}
\usepackage{epsfig}
\usepackage{graphicx}
\usepackage{amsmath}
\usepackage{amssymb}
\usepackage{float}
\usepackage{dblfloatfix}
\graphicspath{ {./images/} }

\usepackage[pagebackref=true,breaklinks=true,letterpaper=true,colorlinks,bookmarks=false]{hyperref}

\cvprfinalcopy 

\def\cvprPaperID{****} 

\ifcvprfinal\fi
\begin{document}

\title{Joint Moment Retrieval and Highlight Detection Via Natural Language Queries}

\author{
Richard Luo\\
{\tt\small rluo47@gatech.edu}
\and
Austin Peng\\
{\tt\small apeng39@gatech.edu}
\and
Koby Beard\\
{\tt\small kbeard8@gatech.edu}
\and
Heidi Yap\\
{\tt\small hyap7@gatech.edu}
\and
Georgia Institute of Technology\\
Atlanta, GA 30332\\
}


\maketitle

\begin{abstract}
    Video summarization has become an increasingly important task in the field of computer vision due to the vast amount of video content available on the internet. In this project, we propose a new method for natural language query based joint video summarization and highlight detection using multi-modal transformers. This approach will use both visual and audio cues to match a user's natural language query to retrieve the most relevant and interesting moments from a video. Our approach employs multiple recent techniques used in Vision Transformers (ViTs) to create a transformer-like encoder-decoder model. We evaluated our approach on multiple datasets such as YouTube Highlights and TVSum to demonstrate the flexibility of our proposed method. Our code and raw results is publicly available at \href{https://github.com/Skyline-9/Visionary-Vids}{https://github.com/Skyline-9/Visionary-Vids}
    
\end{abstract}

\section{Introduction/Background/Motivation}

\subsection{Introduction/Background}
Video content has proliferated in the last few years, now taking up the majority of all internet traffic. However, video is an inherently dense data medium, with a single video encoding large amounts of information about visual, audio, and linguistic elements. With increasingly large amounts of video available, there has been an increasing focus on two important needs: finding a specific, relevant moment in a video and quickly skimming over the a video to summarize its content. This reflects a shift towards summarizing or extracting just the highlights of a video. Twitch, for example, has clips where users can splice just an interesting section of hours-long live streams. Similarly, TikTok, Instagram Reels, and YouTube Shorts all move towards short-form content.

\begin{table*}
\begin{center}

\begin{tabular}{@{}lllll@{}}
\toprule
\multicolumn{1}{c}{\multirow{2}{*}{\textbf{Method}}} & \multicolumn{1}{c}{\textbf{YT}} & \multicolumn{1}{c}{\textbf{TVSum}} & \multicolumn{2}{c}{\textbf{QVHighlights}} \\ 
\cmidrule(l){2-5} 
 & \multicolumn{1}{c}{mAP} & \multicolumn{1}{c}{Top-5 mAP} & MR (mAP) & HD (mAP) \\ 
\cmidrule(l){2-5} 
MINI-Net \cite{hong2020mini} & 64.36 & 73.24 & – & – \\
Joint-VA \cite{badamdorj2021joint} & 71.80 & 76.30 & – & – \\
Moment-DETR \cite{lei2021qvhighlights} & – & – & 34.05 & 37.67 \\
UMT \cite{liu2022umt} & 74.93 & 83.14 & 38.59* & 39.85* \\
\textbf{VisionaryVid (Ours)} & \textbf{76.15} & \textbf{85.50} & 38.28* & 39.12* \\ 
\multicolumn{5}{c}{* denotes close performance to state of the art ($\triangle$ mAP $<$ 1)} \\
\bottomrule
\end{tabular}

\end{center}
\caption{Effectiveness of multimodal learning on YouTube Highlights, TVSum, and QVHighlights}
\label{tab:res1}
\end{table*}

\begin{table*}
    \begin{center}
        \begin{tabular}{@{}ccccc@{}}
            \toprule
            \multirow{3}{*}{\textbf{Method}} & \multicolumn{4}{c}{\textbf{Charades-STA}}     \\ \cmidrule(l){2-5} 
                                             & MR (mAP) & MR (mAP) & HD (mAP) & HD (mAP) \\
                                             & R1@0.5   & R1@0.7   & R5@0.5   & R5@0.7   \\ \cmidrule(r){1-1}
            SAP \cite{chen2019semantic}                              & 27.42    & 13.36    & 66.37    & 38.15    \\
            SM-RL \cite{wang2019language}                            & 24.36    & 11.17    & 61.25    & 32.08    \\
            UMT$^\dagger$ \cite{liu2022umt}                              & \textbf{48.31}    & \textbf{29.25}    & \textbf{88.79}    & \textbf{56.08}    \\
            \textbf{VisionaryVid}$^\dagger$ (Ours)                    & 25.97    & 13.25    & 84.87    & 40.38    \\ \bottomrule
            \multicolumn{5}{c}{$^\dagger$ denotes trained with video + audio}\\
        \end{tabular}
    
    \end{center}
    
    \caption{Effectiveness of multimodal learning on Charades-STA}
    \label{tab:res2}
\end{table*}

\begin{figure*}[t]
    \begin{center}
       \includegraphics[width=150mm]{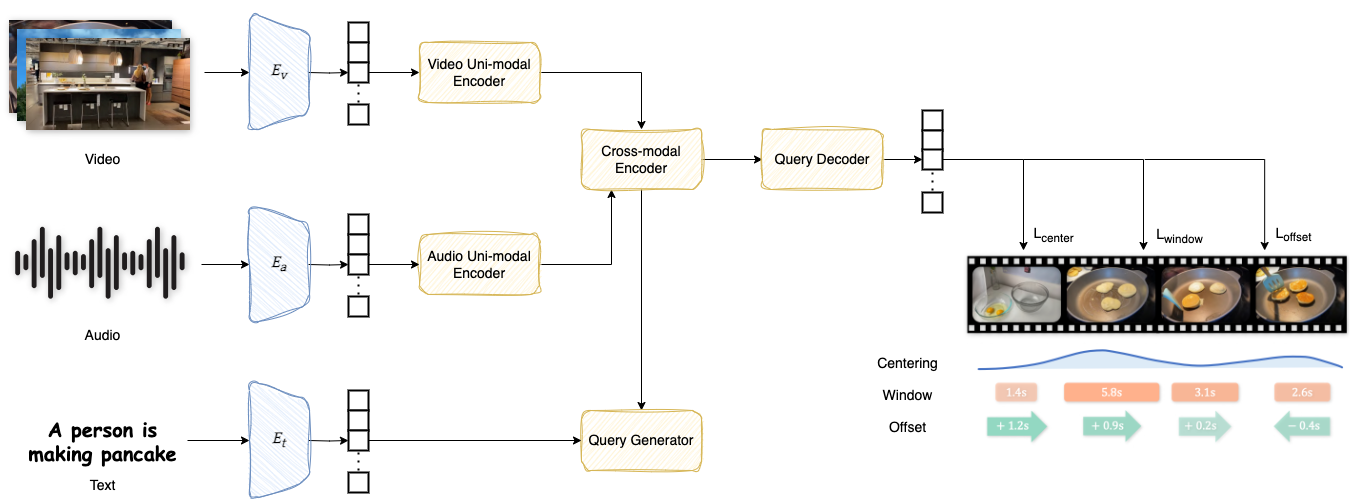}\\
    \end{center}
   \caption{VisionaryVids Model Architecture}
\label{fig:vv-model-arch}
\end{figure*}

\begin{figure*}[t]
    \begin{center}
       \includegraphics[width=150mm]{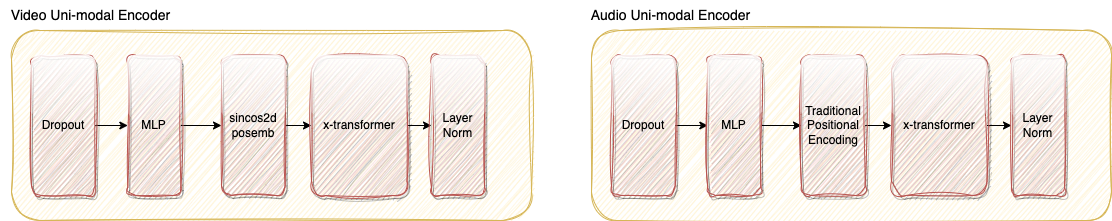}\\
    \end{center}
   \caption{VisionaryVids Video and Audio Encoder}
\label{fig:vv-encoder-arch}
\end{figure*}

\subsection{Related Works}
Currently, video highlighting/summarization is largely done by hand. Vloggers or other content creators need to sift through long streams of video and cut them out to reveal only the most relevant parts. There are many online platforms that crowdsource the process of generating highlights, such as the subreddit r/nba, where users will clip a stream and post it on the forum \cite{rnba}. Highlight generation is also done via machine learning, although many of the approaches are domain specific. For example, various deep learning schemes have been approached to perform automatic highlight detection in soccer \cite{darapaneni2022detecting} \cite{karimi2021soccer}. More recently, there have been attempts to generalize video summarization and highlight detection \cite{rochan2020adaptive}. Another example is video thumbnail generation, which focuses on the downstream task of video summarization by selecting a short video clip to serve as the thumbnail \cite{Xu_Bai_Shi_Chen_Gao_Tian_Zhou_Sun_2021}. However, video is an inherently data rich medium, as it contains information about a myriad of things such as objects, language, movement, sound, and time. This brings about a need for multi-modal models. There have been some attempts at multi-modal models, but they lack the critical ability to query specific moments \cite{takahashi2017aenet}. In video thumbnail generation, which relies on video highlight summarization, early approaches used graph convolutional networks to model interactions between the word and text queries \cite{rochan2020thumbnail}. Other approaches that have the ability to query moments and generate highlights lack the information gained in a multi-modal model \cite{lin2022videogenic} \cite{lei2021qvhighlights}.

\subsection{Motivation}

Video summarization and highlight detection have become increasingly important tasks in the field of computer vision, with significant potential to benefit a broad range of stakeholders, including content creators, marketers, researchers, and end-users. Content creators and marketers can use video summarization and highlight detection to increase engagement and retention for their target audience. By identifying and highlighting the most relevant and interesting parts of their video content, they can keep viewers engaged and motivated to watch more, leading to higher viewership, increased brand exposure, and potentially, higher revenues. On the other hand, researchers can use video summarization and highlight detection to improve the efficiency and effectiveness of data analysis, enabling them to quickly identify the most relevant and interesting parts of their videos for further investigation. This can help accelerate research progress and lead to new insights in various fields, such as social sciences, psychology, and medicine.

End-users can benefit from video summarization and highlight detection by being able to quickly find the most relevant and interesting parts of a video, saving them time and effort. This is especially important with the vast amount of video content available on the internet nowadays, where users are usually short on time and attention.

If successful, our proposed method using multi-modal transformers for natural language query-based joint video summarization and highlight detection has the potential to significantly improve the efficiency and effectiveness of video summarization and highlight detection, benefiting all of these stakeholders. This has the opportunity to ultimately lead to a more engaging, informative, and enjoyable video experience for all users.

\subsection{Datasets}

{\bf QVHighlights \cite{lei2021qvhighlights}:}

QVHighlights is a large-scale video dataset that contains 4,000 online video clips from 35 popular sports categories, including basketball, soccer, tennis, and more. The dataset includes manually annotated highlight timestamps and corresponding importance scores from 3-5 annotators for each video clip. In addition, it also contains audio transcripts and visual frames extracted from each video clip. This dataset was introduced in the paper "QVHighlights: A Large-Scale Video Dataset for Highlight Detection in Sports" by Lei et al. in 2021.\\

{\bf Charades-STA \cite{GaoTemporal}:}

Charades-STA is a video captioning and temporal localization dataset that includes 9,848 short video clips of daily activities, such as brushing teeth, cooking, and more. The dataset provides 33,753 natural language sentences for describing the video content, as well as temporal annotations for the beginning and ending of the action segments within each video clip. Charades-STA also includes a challenge set of 500 videos with more complex activities and multiple action segments. This dataset was introduced in the paper "Charades-STA: A Benchmark Dataset for Natural Language Visual Reasoning and Scripting" by Gao et al. in 2017.\\

{\bf YouTube Highlights \cite{sun2014ranking}:}

YouTube Highlights is a video highlight dataset that contains 12,000 video clips from 20 popular sports categories, including basketball, soccer, and football. Each video clip is annotated with the start and end timestamps of the highlights and the corresponding importance score from 5 annotators. The dataset also includes audio transcripts and visual frames extracted from each video clip. This dataset was introduced in the paper "Ranking and Classifying Atypical Activities in Video" by Sun et al. in 2014.\\

{\bf TVSum \cite{song2015tvsum}:}

TVSum is a video summarization dataset that contains 50 videos of various topics, such as news, documentaries, and sports. The dataset provides annotations of the keyshot frames and their corresponding importance scores, which are based on the amount of information presented in the video. In addition, TVSum also includes annotations of the human-created summaries for each video, as well as baseline summaries generated by different summarization methods. This dataset was introduced in the paper "TVSum: Summarizing web videos using titles" by Song et al. in 2015.

\section{Approach}

\begin{table*}
    \begin{center}
        \begin{tabular}{@{}lccccccccccl@{}}
            \toprule
            \multicolumn{1}{c}{\multirow{3}{*}{\textbf{Method}}} & \multicolumn{11}{c}{\textbf{TVSum}} \\ \cmidrule(l){2-12} 
            \multicolumn{1}{c}{} & \multicolumn{11}{c}{HD (Top-5 mAP)} \\ \cmidrule(l){2-12} 
            \multicolumn{1}{c}{} & BK & BT & DS & FM & GA & MS & PK & PR & VT & VU & Avg. \\
            \midrule
            MINI-Net \cite{hong2020mini} & 75.0 & 80.2 & 65.5 & 57.8 & 78.2 & 81.8 & 78.1 & 65.8 & 80.6 & 68.3 & 73.1 \\
            Joint-VA \cite{badamdorj2021joint} & \multicolumn{1}{l}{73.0} & \multicolumn{1}{l}{\textbf{97.4}} & \multicolumn{1}{l}{67.5} & \multicolumn{1}{l}{70.0} & \multicolumn{1}{l}{78.5} & \multicolumn{1}{l}{86.1} & \multicolumn{1}{l}{80.1} & \multicolumn{1}{l}{69.2} & \multicolumn{1}{l}{83.7} & \multicolumn{1}{l}{57.3} & 76.3 \\
            TCG \cite{ye2021temporal} & 77.3 & 78.6 & 68.1 & 71.6 & 81.9 & 78.6 & 80.2 & 75.5 & 85.0 & 71.4 & 76.8 \\
            UMT \cite{liu2022umt} & 86.9 & 84.4 & 79.6 & 76.0 & 88.2 & 78.8 & 81.4 & \textbf{87.0} & \textbf{87.5} & 81.5 & 83.1 \\
            \textbf{VisionaryVids} (Ours) & \textbf{92.4} & 86.5 & \textbf{80.3} & \textbf{79.4} & \textbf{91.7} & \textbf{87.3} & \textbf{82.9} & 82.7 & 84.9 & \textbf{87.1} & \textbf{85.5} \\
            \bottomrule
        \end{tabular}

    \end{center}

    \caption{Effectiveness of multimodal learning on TVSum. Comparison with representative highlight detection methods on TVSum dataset. Above are the methods using visual-audio features.}
    \label{tab:res3}
\end{table*}

\begin{table*}
    \begin{center}
        \centering
        \begin{tabular}{@{}lccccccc@{}}
            \toprule
            \multicolumn{1}{c}{\multirow{3}{*}{\textbf{Method}}} & \multicolumn{7}{c}{\textbf{YT}} \\
            \cmidrule(l){2-8} 
            \multicolumn{1}{c}{} & \multicolumn{7}{c}{HD (mAP)} \\
            \cmidrule(l){2-8} 
            \multicolumn{1}{c}{} & Dog & Gym. & Par. & Ska. & Ski. & Sur. & Avg. \\
            \midrule
            MINI-Net \cite{hong2020mini} & 58.2 & 61.7 & 70.2 & 72.2 & 58.7 & 65.1 & 64.4 \\
            Joint-VA \cite{badamdorj2021joint} & 55.4 & 62.7 & 70.9 & 69.1 & 60.1 & 59.8 & 63.0 \\
            TCG \cite{ye2021temporal} & 64.5 & 71.9 & 80.8* & 62.0 & \textbf{73.2} & 78.3 & 71.8 \\
            UMT \cite{liu2022umt} & 65.9 & 75.2* & \textbf{81.6} & 71.8 & 72.3* & \textbf{82.7*} & 74.9 \\
            VisionaryVids (Ours) & \textbf{70.0} & \textbf{75.2*} & 78.2 & \textbf{79.4} & 72.0 & 82.1* & \textbf{76.2} \\
            \multicolumn{8}{c}{* denotes close performance to state of the art ($\triangle$ mAP $<$ 1)}\\
            \bottomrule
        \end{tabular}
    \end{center}

    \caption{Experimental results on YouTube Highlights dataset. Above are the methods using visual-audio features.}
    \label{tab:res4}
\end{table*}

\subsection{Overview}
Our team was inspired by the work done by Unified Multi-modal Transformers \cite{liu2022umt}, vision transformers (ViT) \cite{dosovitskiy2021vit}, and the improved vision transformers \cite{beyer2022bettervit} and combined these state of the art efforts for our model's novelty. Our model is shown in Figure \ref{fig:vv-model-arch} and Figure \ref{fig:vv-encoder-arch}.

In the UMT model, we decided to improve the uni-modal encoder for both video and audio. For video, we took inspiration from improvements  made to the vanilla vision transformer, specifically in the positional encoding. The original transformer paper uses a sinusoidal positional embedding, which is used in the original UMT paper \cite{vaswani2017attention}. However, recent advances to ViTs found that using a sine-cosine 2D positional encoding achieved better results by capturing more fine-grained local details \cite{beyer2022bettervit}.

Additionally, for both the video and audio uni-modal encoders, we improved the standard transformer self-attention mechanism by augmenting with persistent memory. The idea is to combine the self-attention layer and the feed-forward layer. This would simplify the model with no performance loss while also being capable of storing contextual information. However, empirical results have shown that including both the feed-forward layer and persistent memory yield better results \cite{persistentmemory}.

Finally, we brought in a relatively new, simple optimizer called Lion (EvoLved Sign Momentum). Lion caught our attention as it brought ViT to the top performer on the ImageNet benchmark. It is memory-efficient and achieves stronger generalizations across architectures when compared to optimizers (ex. Adam, AdamW, Adafactor) used in recent state-of-the-art models.

\subsection{Positional Encoder}
The original ViT paper \cite{dosovitskiy2021vit} uses standard 1d positional embeddings. The improved vision paper transformers paper \cite{beyer2022bettervit}  uses 2d positional embdeddings for better results. Specifically, a 2d sin cos function is used for positional embedding. The 2d function allows for improved representation of the spatial information of the input. We use this same 2d sin cos positional embedding. 

\subsection{Persistent Memory}
The idea of persistent memory was first introduced as a replacement for the feed-forward layer \cite{persistentmemory}. The persistent memory block allows the self-attention layer and feedforward layer to be combined into one, simplifying the model. Persistent memory works by concatenating the weights of the feedforward vectors onto the key and value vectors of the self-attention layer. Empirical findings suggest that keeping the feedforward layer while adding persistent memory leads to even better performance \cite{x-transformers}. Our model leverages this optimization, and uses the self-attention layer with persistent memory and a feedforward layer. 

\subsection{Lion Optimizer}
The Lion optimizer is a recently proposed optimizer \cite{lionoptimizer} that was discovered from a program search for optmization algorithms for deep neural networks. To close the gap between proxy and target tasks, Lion introduces a program selection and simplification strategy using evolved sign momentum.

Some benefits of Lion are that is is more memory efficient than Adam because it only tracks momentum. In addition, it uses the same magnitude for the update unlike adaptive optimizers. The main difference is that it uses the sign operation to calculate the magnitude.


The Adam optimizer is a widely used optimization algorithm that is based on adaptive moment estimation. The optimizer computes adaptive learning rates for each parameter based on the first and second moments of the gradients. This allows the optimizer to converge more quickly and efficiently than traditional gradient descent algorithms. It also incorporates momentum, which helps the optimizer to continue moving in the same direction when the gradients are consistent. This can help to speed up convergence and avoid getting stuck in local optima.

One paper \cite{lionoptimizer} presents Lion and evaluates its performance against widely used optimizers such as Adam and Adafactor on various deep learning models and tasks. The experiments show that Lion outperforms Adam and Adafactor in several scenarios; Lion boosts the accuracy of ViT (Vision Transformer) by up to 2\% on ImageNet and reduces pre-training compute by up to 5x on JFT (JFT-300M, a large-scale image-text dataset). In vision-language contrastive learning, Lion achieves 88.3\% and 91.1\% accuracy on ImageNet, surpassing the previous state-of-the-art results by 2\% and 0.1\%, respectively. Lion also outperforms Adam in diffusion models by achieving a better FID (Fréchet Inception Distance) score and reducing training compute by up to 2.3x. For autoregressive, masked language modeling, and fine-tuning, LION exhibits similar or better performance compared to Adam.

In this study, we investigate the performance of the Lion optimizer as an alternative to the widely used Adam optimizer for training the UMT model by implementing Lion and integrating it into the existing code. 

\subsection{Problems}
Initially, we anticipated extremely long training times due to the large structure of the model and the many many transformers used. However, after using PyTorch 2's torch.compile optimization, we were able to significantly speed up our training speed. Additionally, we gained access to a multi-GPU server, which allowed us to train the larger datasets such as QVHighlights and Charades-STA.

Our first approach was actually to use convolutions (e.g. ConvNeXt v2) as an encoder layer instead of the typical TransformerEncoderLayer. This idea was that convolutions are inherently translation invariant and locality sensitive, which allows them to perform better when not a lot of data is fed in. Because our datasets are relatively small compared to something like ImageNet (which has over 14 million images), the presupposition was that convolutions would outperform transformers. However, because we are using feature extractors before feeding in the code, the convolutions would not have the raw image data fed into them, which makes them lose many of their inherent values. As such, we pivoted towards improving the transformer architectures.

\section{Experiments and Results}

\subsection{Experimental Settings}

\vspace{1em}
\noindent\textbf{Evaluation Metrics}\hspace{1em} Following existing works, we use the same evaluation metrics based off mean average precision (mAP) \cite{liu2022umt}. In the YouTube Highlights, mAP is used, and in the TV-Sum dataset, top 5 mAP is used. For QVHighlights moment retrieval, we adopt Recall@1 with IoU thresholds of 0.5 and 0.7, mAP with IoU thresholds 0.5 and 0.75, and average mAP over a series of IoU thresholds [0.5:0.05:0.95]. For QVHighlights highlight detection, we adopt mAP
and HIT@1, where a clip prediction is treated as a true positive if it has the saliency score of Very Good. In the CharadesSTA dataset, we used Recall@1 and Recall@5 with IoU thresholds 0.5 and 0.7.

\vspace{1em}
\noindent\textbf{Feature Extraction}\hspace{1em}

Given an untrimmed video and a query, our research aims to find all the video moments where visual-audio contents are relevant to the query. To simplify our task, we utilize pre-trained feature extractors to extract visual, audio, and textual features from video and clip datasets.

More specifically, with QVHighlights, we leveraged the extracted features using SlowFast \cite{feichtenhofer2019slowfast} and CLIP \cite{radford2021learning}. For Charades-STA, we utilized optical flow features, VGG \cite{simonyan2015deep}, and GloVe \cite{pennington2014glove} embeddings.

For the remaining two datasets, TVSum and YouTube Highlights, we leveraged I3D \cite{carreira2018quo} pre-trained on Kinetics 400 \cite{kay2017kinetics} to obtain visual features. For TVSum, we also extracted the titles using CLIP.

For all four datasets, we extracted audio features using a PANN \cite{kong2020panns} model pre-trained on AudioSet \cite{gemmeke2018audioset}. The corresponding visual and audio features of each short segment of the video (aka clip) was synchronized to occur at the same time interval using timestamps.

Note: The extracted video and audio features are fed into separate uni-modal encoders then fused by a cross-modal encoder, following the method from UMT \cite{liu2022umt}.

\vspace{1em}
\noindent\textbf{Model Configuration}\hspace{1em}

In YouTube Highlights and TVSum, we use 1 decoder layer each, and in QVHighlights and Charades-STA, we use 3 decoder layers due to the larger dataset size. In each of the experiments, we use only one cross-modal encoder and one uni-modal encoder each for visual features, audio features, and query. Following UMT \cite{liu2022umt}, we use learning sincos2d positional encodings, pre-norm style layer normalization, 8 attention heads, and a dropout rate of 0.1 for all the transformers, with extra pre-dropouts with rate 0.5 for visual and audio inputs, and 0.3 for text inputs. As for the optimizer, we used the Lion optimizer \cite{lionoptimizer, lion-pytorch}. On YouTube Highlights and TVSum, we used learning rate 1e-4, 5e-4, and 1e-3 with weight decay 1e-4. We evaluated each dataset on the best combination of each of these and took the best result. For Charades and QVHighlights, the training time was much longer, so we only tested one combination of learning rate 1e-4 and weight decay 1e-4. The training of the model involved using various batch sizes and epoch numbers for different datasets. Specifically, a batch size of 32 was used for 200 epochs on QVHighlights, a batch size of 8 was used for 100 epochs on Charades-STA, a batch size of 4 was used for 100 epochs on YouTube Highlights, and a batch size of 1 was used for 500 epochs on TVSum.

\subsection{VisionaryVids Joint Moment Retrieval and Highlight Detection Results}

We first evaluate our model on the YouTube Highlights, TVSum, and QVHighlights dataset shown in Table \ref{tab:res1}, in comparison to all the other performances reported before. On highlight detection, we outperform previous state-of-the-art method UMT in the YouTube Highlights and TVSum dataset. However, on QVHighlights, we perform near state-of-the-art in both moment retrieval and highlight detection. This is likely due to the lack of hyperparameter tuning for QVHighlights, as the larger dataset and more complicated data required more detail. Specifically, it's important to note that both our approach with VisionaryVids and UMT use more decoder layers in QVHighlights than in YouTube and TVSum due to the larger scale of the dataset, and this may be a factor to tune as a hyperparameter. Since the loss curve for QVHighlights in Figure \ref{fig:qvcurve} ends up going back up, we believe that our hyperparameteres were faulty. With sufficient time and resources to perform hyperparameter tuning on QVHighlights, we can improve the performance of VisionaryVids.
    
Next, we evaluated our model on Charades-STA against UMT and other pre-existing methods using video and audio. While UMT also used the optical flow features of Charades instead of audio, we did not have time to test the optical flow features \cite{liu2022umt}. These results are shown in Table \ref{tab:res2}. Surprisingly, our model performed terribly compared to UMT on moment retrieval, and slightly worse on highlight detection. Without more hyperparameter tuning, it's difficult to say whether our model architecture has large flaws or if the hyperparameters we chose as default performed extremely poorly. Another key characteristic to note is that Charades-STA is a large dataset, even larger than QVHighlights.

\subsection{Strengths/Weaknesses of Our Approach}

First, looking at the TVSum results in Table \ref{tab:res3}, our model has a higher ceiling than previous state-of-the-art UMT. On the BK dataset in TVSum, our model performed over 5\% better as measured by mAP.  Similarly, we outperform UMT again by over 5\% on the VU dataset. This trend holds again for the YouTube Highlights results in Table \ref{tab:res4}. We outperform previous state-of-the-art UMT again on Dog by over 4\% and over 7\% on Skate.

\begin{figure}[t]
\begin{center}
   \includegraphics[width=75mm,scale=0.5]{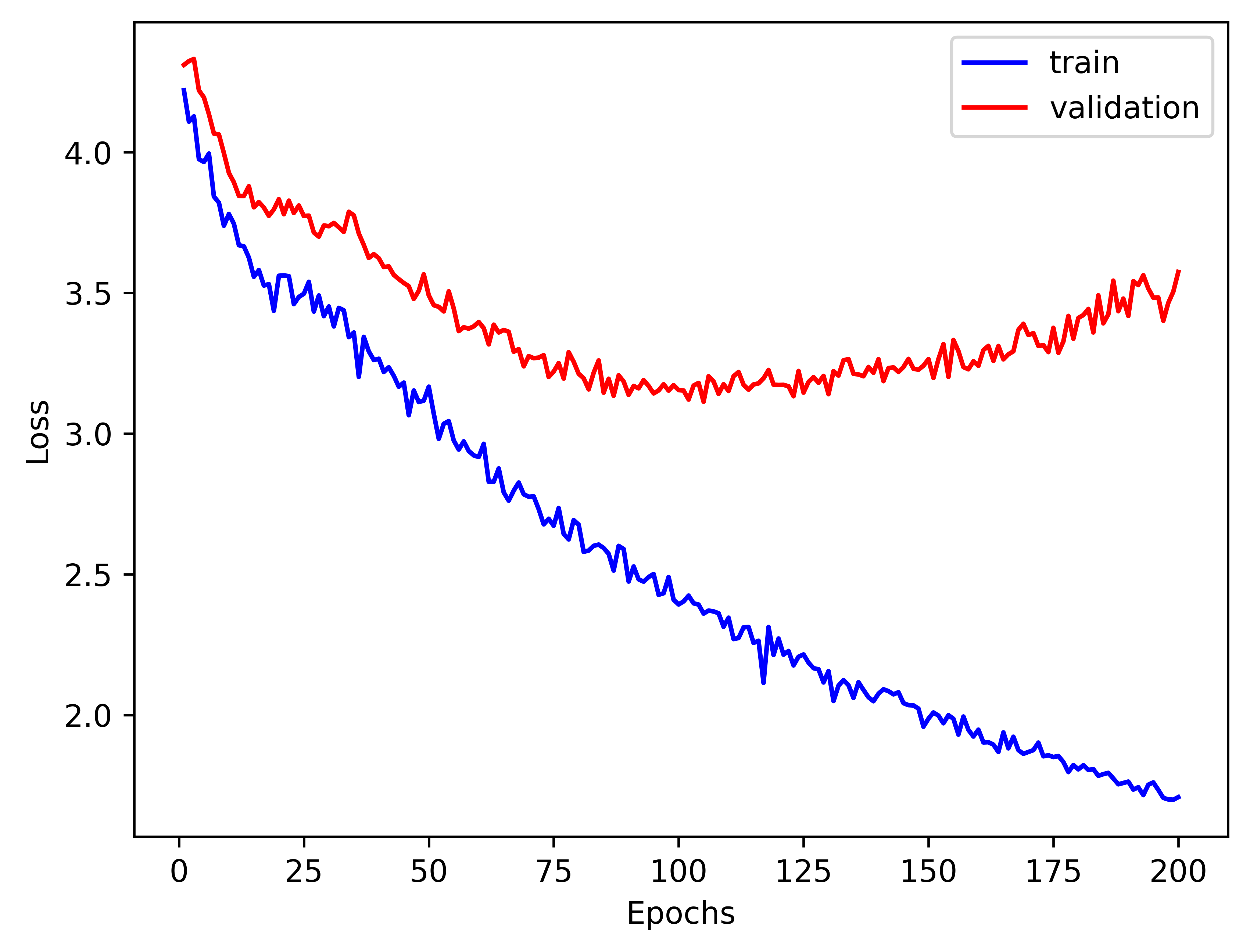}\\
\end{center}
   \caption{Loss Curve for QVHighlights}
\label{fig:qvcurve}
\end{figure}

\begin{figure}[H]
    \begin{center}
        \includegraphics[width=75mm,scale=0.5]{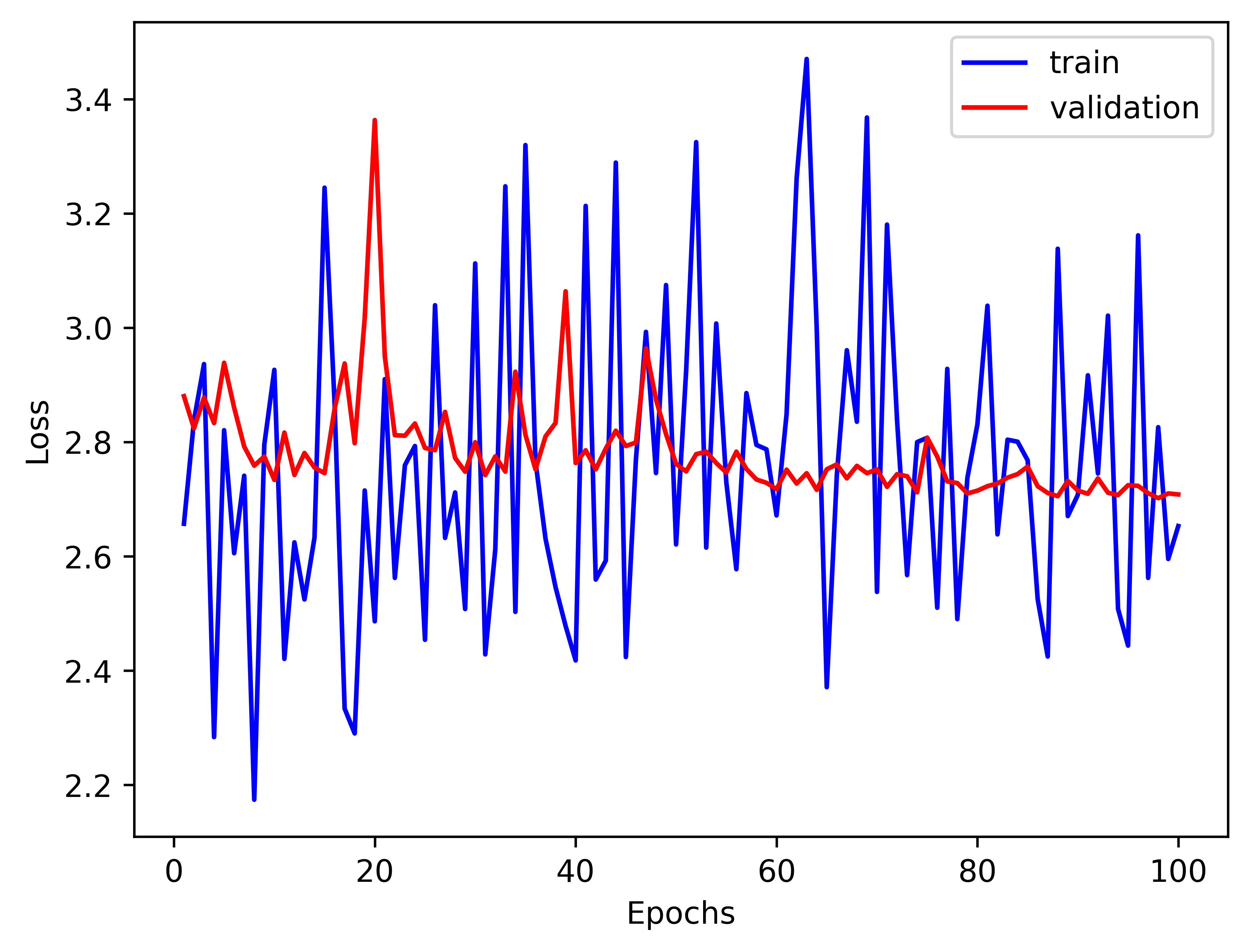}\
    \end{center}
    \caption{Loss Curve for Charades-VA}
    \label{fig:charadescurve}
\end{figure}

However, we perform similarly to state-of-the-art on QVHighlights and worse on Charades-STA. Looking at the loss curves in Figure \ref{fig:qvcurve}, we can see that we are drastically overfitting the training set. Quantitatively, our validation loss is twice as high as our training loss! Future research can be done investigating how to reduce overfitting on the QVHighlights dataset, either using increased weight decay, different batch sizes, increasing dropout in the transformers, or changing the number of decoder layers. Looking at the loss curves in Figure \ref{fig:charadescurve}, we see that loss is not decreasing after just a few epochs, and even worse, the training loss is very erratic and high. This likely suggests underfitting, and possibly also a bad set of hyperparameters, as we are stuck in a local optima.

\begin{table*}
    \begin{center}
    \begin{tabular}{|l|c|p{8cm}|}
    \hline
    Student Name & Contributed Aspects & Details \\
    \hline\hline
    Richard Luo & Data Collection, Implementation, Training & Wrote scripts to download all four datasets, implemented sincos2D positional embedding, implemented persistent memory in transformer layer, and trained the model. \\
    Austin Peng & Implementation, Data Visualization & Trained/validated the model for the YouTube Highlights dataset. Compared model output with \cite{liu2022umt} through plot visualizations. \\
    Heidi Yap & Lion Implementation and Data Visualization & Helped Koby with implementing the Lion optimizer. Then trained/validated model for TVSum dataset and wrote the script for visualizing the loss curve for all of the metrics JSON files. \\
    Koby Beard & Lion Implementation and Training & Implemented the Lion optimizer. Experimented with Lion vs. Adam optimizer. Also trained/validated the model for the TVSum dataset. \\

    \hline
    \end{tabular}
    \end{center}
    \caption{Contributions of team members.}
    \label{tab:contributions}
\end{table*}

\section{Work Division}

Our work division is shown in Table \ref{tab:contributions}.

{\small
\bibliographystyle{ieee_fullname}
\bibliography{egbib}

\begin{thebibliography}{10}\itemsep=-1pt

\bibitem{rnba}
r/nba subreddit.
\newblock Accessed: 2023-04-26.

\bibitem{badamdorj2021joint}
Taivanbat Badamdorj, Mrigank Rochan, Yang Wang, and Li Cheng.
\newblock Joint visual and audio learning for video highlight detection.
\newblock In {\em Proceedings of the IEEE/CVF International Conference on
  Computer Vision (ICCV)}, pages 8127--8137, 2021.

\bibitem{beyer2022bettervit}
Lucas Beyer, Xiaohua Zhai, and Alexander Kolesnikov.
\newblock Better plain vit baselines for imagenet-1k, 2022.

\bibitem{carreira2018quo}
Joao Carreira and Andrew Zisserman.
\newblock Quo vadis, action recognition? a new model and the kinetics dataset,
  2018.

\bibitem{chen2019semantic}
Shaoxiang Chen and Yu-Gang Jiang.
\newblock Semantic proposal for activity localization in videos via sentence
  query.
\newblock In {\em Proceedings of the AAAI Conference on Artificial Intelligence
  (AAAI)}, pages 8199--8206, 2019.

\bibitem{lionoptimizer}
Xiangning Chen, Chen Liang, Da Huang, Esteban Real, Kaiyuan Wang, Yao Liu, Hieu
  Pham, Xuanyi Dong, Thang Luong, Cho-Jui Hsieh, Yifend Lu, and Quoc~V. Le.
\newblock Symbolic discovery of optimization algorithms, 2023.

\bibitem{darapaneni2022detecting}
Narayana Darapaneni, PrashanthAnand Kumar, Nikhil Malhotra, Vigneswaran
  Sundaramurthy, Abhaya Thakur, Shivam Chauhan, Krishna~Chaitanya Thangeda, and
  Anwesh~Reddy Paduri.
\newblock Detecting key soccer match events to create highlights using computer
  vision.
\newblock {\em ArXiv}, abs/2204.02573, 2022.

\bibitem{dosovitskiy2021vit}
Alexey Dosovitskiy, Lucas Beyer, Alexander Kolesnikov, Dirk Weissenborn,
  Xiaohua Zhai, Thomas Unterthiner, Mostafa Dehghani, Matthias Minderer, Georg
  Heigold, Sylvain Gelly, Jakob Uszkoreit, and Neil Houlsby.
\newblock An image is worth 16x16 words: Transformers for image recognition at
  scale, 2021.

\bibitem{feichtenhofer2019slowfast}
Christoph Feichtenhofer, Haoqi Fan, Jitendra Malik, and Kaiming He.
\newblock Slowfast networks for video recognition.
\newblock In {\em Proceedings of the IEEE/CVF International Conference on
  Computer Vision (ICCV)}, October 2019.

\bibitem{GaoTemporal}
Jiyang Gao, Chen Sun, Zhenheng Yang, and Ram Nevatia.
\newblock Tall: Temporal activity localization via language query.
\newblock In {\em Proceedings of the IEEE/CVF International Conference on
  Computer Vision (ICCV)}, page 5267–5275, 2021.

\bibitem{gemmeke2018audioset}
Jort~F. Gemmeke, Daniel P.~W. Ellis, Dylan Freedman, Aren Jansen, Wade
  Lawrence, R.~Channing Moore, Manoj Plakal, and Marvin Ritter.
\newblock Audio set: An ontology and human-labeled dataset for audio events.
\newblock In {\em 2017 IEEE International Conference on Acoustics, Speech and
  Signal Processing (ICASSP)}, pages 776--780, 2017.

\bibitem{hong2020mini}
Fa-Ting Hong, Xuanteng Huang, Wei-Hong Li, and Wei-Shi Zheng.
\newblock Mini-net: Multiple instance ranking network for video highlight
  detection.
\newblock In {\em Proceedings of the European Conference on Computer Vision
  (ECCV)}, pages 345--360, 2020.

\bibitem{karimi2021soccer}
Ali Karimi, Ramin Toosi, and Mohammad~Ali Akhaee.
\newblock Soccer event detection using deep learning, 2021.

\bibitem{kay2017kinetics}
Will Kay, Joao Carreira, Karen Simonyan, Brian Zhang, Chloe Hillier, Sudheendra
  Vijayanarasimhan, Fabio Viola, Tim Green, Trevor Back, Paul Natsev, Mustafa
  Suleyman, and Andrew Zisserman.
\newblock The kinetics human action video dataset, 2017.

\bibitem{kong2020panns}
Qiuqiang Kong, Yin Cao, Turab Iqbal, Yuxuan Wang, Wenwu Wang, and Mark~D.
  Plumbley.
\newblock Panns: Large-scale pretrained audio neural networks for audio pattern
  recognition, 2020.

\bibitem{lei2021qvhighlights}
Jie Lei, Tamara~L. Berg, and Mohit Bansal.
\newblock Qvhighlights: Detecting moments and highlights in videos via natural
  language queries, 2021.

\bibitem{lin2022videogenic}
David Chuan-En Lin, Fabian~Caba Heilbron, Joon-Young Lee, Oliver Wang, and
  Nikolas Martelaro.
\newblock Videogenic: Video highlights via photogenic moments, 2022.

\bibitem{liu2022umt}
Ye Liu, Siyuan Li, Yang Wu, Chang~Wen Chen, Ying Shan, and Xiaohu Qie.
\newblock Umt: Unified multi-modal transformers for joint video moment
  retrieval and highlight detection.
\newblock In {\em Proceedings of the IEEE/CVF Conference on Computer Vision and
  Pattern Recognition (CVPR)}, pages 3042--3051, 2022.

\bibitem{pennington2014glove}
Jeffrey Pennington, Richard Socher, and Christopher Manning.
\newblock {G}lo{V}e: Global vectors for word representation.
\newblock In {\em Proceedings of the 2014 Conference on Empirical Methods in
  Natural Language Processing ({EMNLP})}, pages 1532--1543, Doha, Qatar, Oct.
  2014. Association for Computational Linguistics.

\bibitem{radford2021learning}
Alec Radford, Jong~Wook Kim, Chris Hallacy, Aditya Ramesh, Gabriel Goh,
  Sandhini Agarwal, Girish Sastry, Amanda Askell, Pamela Mishkin, Jack Clark,
  Gretchen Krueger, and Ilya Sutskever.
\newblock Learning transferable visual models from natural language
  supervision, 2021.

\bibitem{rochan2020thumbnail}
Mrigank Rochan, Mahesh Kumar~Krishna Reddy, and Yang Wang.
\newblock Sentence guided temporal modulation for dynamic video thumbnail
  generation.
\newblock In {\em Proceedings of the British Machine Vision Conference (BMVC)},
  2020.

\bibitem{rochan2020adaptive}
Mrigank Rochan, Mahesh Kumar~Krishna Reddy, Linwei Ye, and Yang Wang.
\newblock Adaptive video highlight detection by learning from user history,
  2020.

\bibitem{simonyan2015deep}
Karen Simonyan and Andrew Zisserman.
\newblock Very deep convolutional networks for large-scale image recognition,
  2015.

\bibitem{song2015tvsum}
Yale Song, Jordi Vallmitjana, Amanda Stent, and Alejandro Jaimes.
\newblock Tvsum: Summarizing web videos using titles.
\newblock In {\em Proceedings of the IEEE/CVF Conference on Computer Vision and
  Pattern Recognition (CVPR)}, pages 5179--5187, 2015.

\bibitem{persistentmemory}
Sainbayar Sukhbaatar, Edouard Grave, Guillaume Lample, Herve Jegou, and Armand
  Joulin.
\newblock Augmenting self-attention with persistent memory, 2019.

\bibitem{sun2014ranking}
Min Sun, Ali Farhadi, and Steve Seitz.
\newblock Ranking domain-specific highlights by analyzing edited videos.
\newblock In {\em Proceedings of the European Conference on Computer Vision
  (ECCV)}, pages 787--802, 2014.

\bibitem{takahashi2017aenet}
Naoya Takahashi, Michael Gygli, and Luc~Van Gool.
\newblock Aenet: Learning deep audio features for video analysis, 2017.

\bibitem{vaswani2017attention}
Ashish Vaswani, Noam Shazeer, Niki Parmar, Jakob Uszkoreit, Llion Jones,
  Aidan~N. Gomez, Lukasz Kaiser, and Illia Polosukhin.
\newblock Attention is all you need, 2017.

\bibitem{lion-pytorch}
Phil Wang.
\newblock Github: lion-pytorch.

\bibitem{x-transformers}
Phil Wang.
\newblock Github: x-transformers.

\bibitem{wang2019language}
Weining Wang, Yan Huang, and Liang Wang.
\newblock Language-driven temporal activity localization: A semantic matching
  reinforcement learning model.
\newblock In {\em Proceedings of the IEEE/CVF Conference on Computer Vision and
  Pattern Recognition (CVPR)}, pages 334--343, 2019.

\bibitem{Xu_Bai_Shi_Chen_Gao_Tian_Zhou_Sun_2021}
Yi Xu, Fan Bai, Yingxuan Shi, Qiuyu Chen, Longwen Gao, Kai Tian, Shuigeng Zhou,
  and Huyang Sun.
\newblock Gif thumbnails: Attract more clicks to your videos.
\newblock {\em Proceedings of the AAAI Conference on Artificial Intelligence},
  35(4):3074--3082, May 2021.

\bibitem{ye2021temporal}
Qinghao Ye, Xiyue Shen, Yuan Gao, Zirui Wang, Qi Bi, Ping Li, and Guang Yang.
\newblock Temporal cue guided video highlight detection with low-rank
  audio-visual fusion.
\newblock In {\em Proceedings of the IEEE/CVF International Conference on
  Computer Vision (ICCV)}, pages 7950--7959, 2021.

\end{thebibliography}
}

\onecolumn
\newpage
\twocolumn

\section*{Appendix}

\subsection{Loss Curves}

\begin{figure}[H]
    \begin{center}
        \includegraphics[width=75mm,scale=0.5]{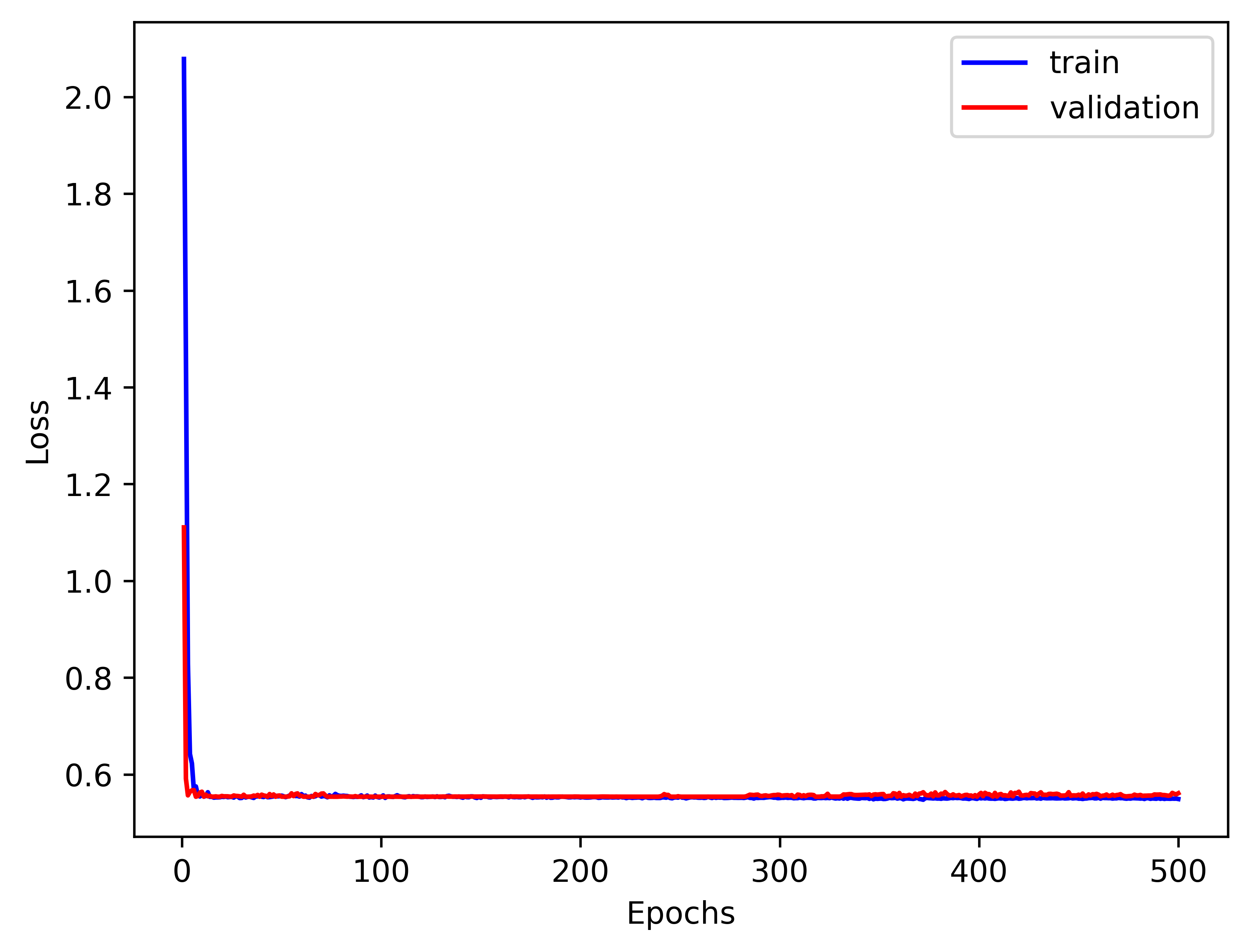}\
    \end{center}
    \caption{Loss Curve for TVSum-BK}
    \label{fig:tvsum-bk}
\end{figure}

\begin{figure}[H]
\centering
\includegraphics[width=75mm,scale=0.5]{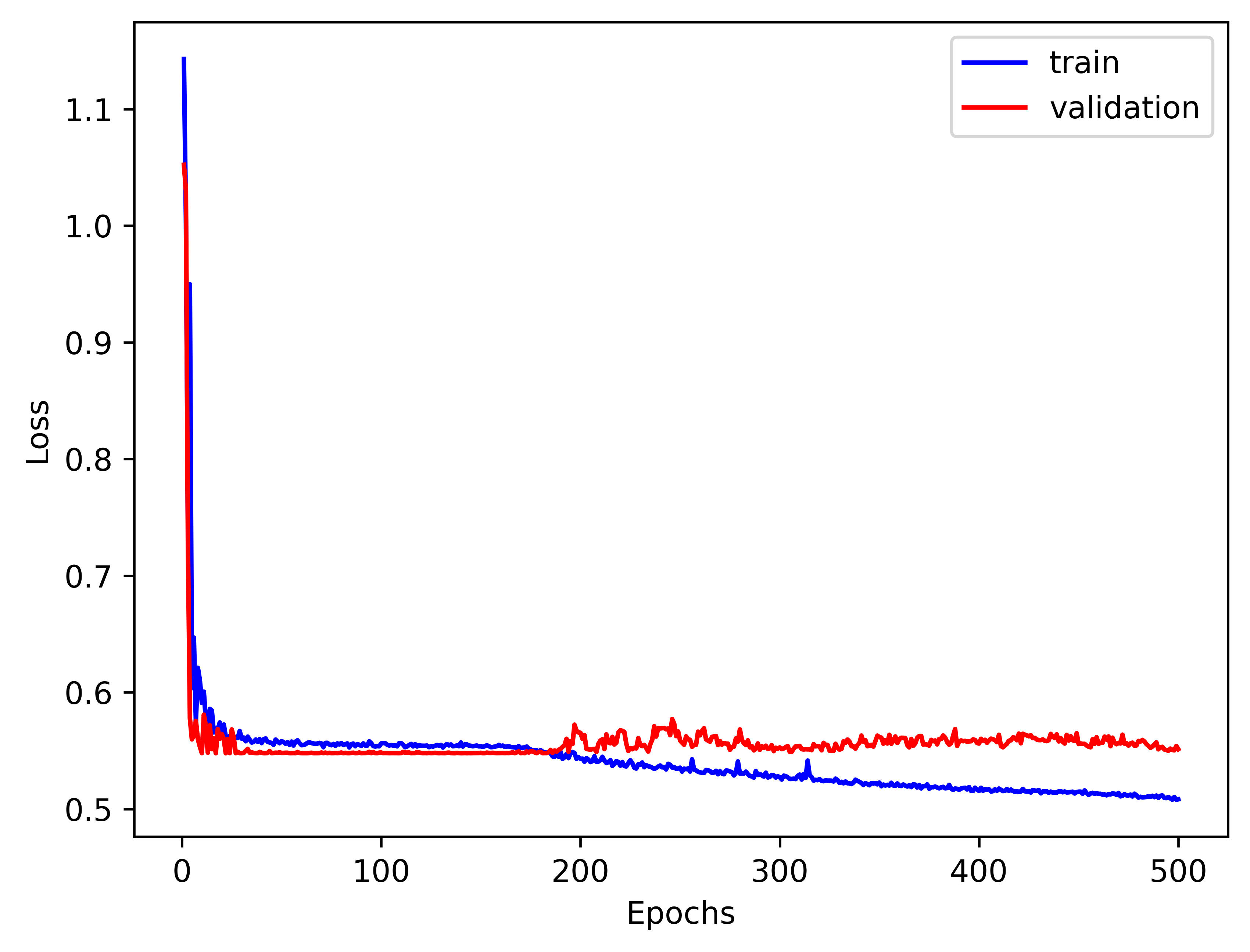}
\caption{Loss Curve for TVSum-BT}
\label{fig:tvsum-bt}
\end{figure}

\begin{figure}[H]
\begin{center}
\includegraphics[width=75mm,scale=0.5]{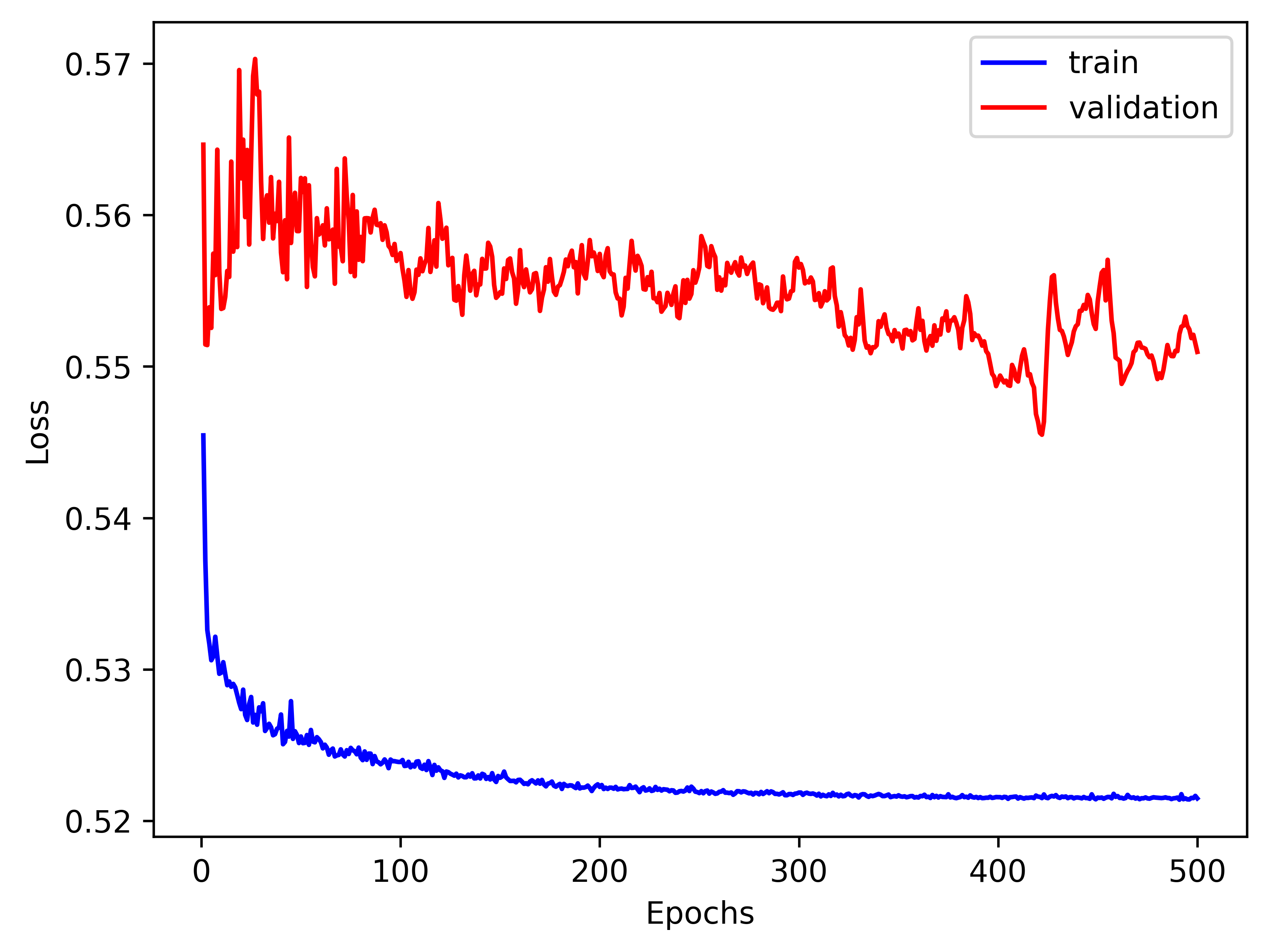}\
\end{center}
\caption{Loss Curve for TVSum-DS}
\label{fig:tvsum-ds}
\end{figure}

\begin{figure}[H]
    \begin{center}
       \includegraphics[width=75mm,scale=0.5]{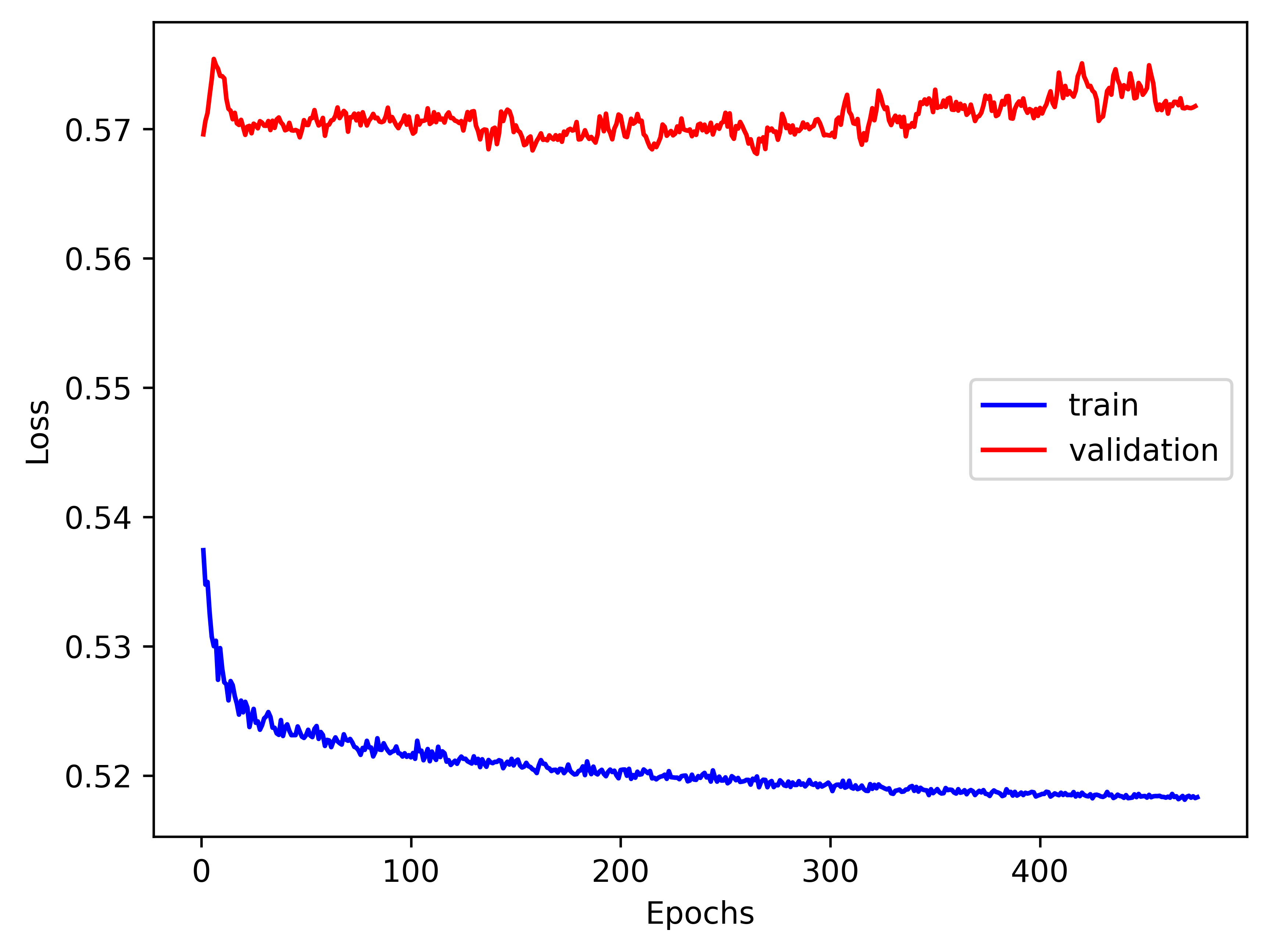}\\
    \end{center}
   \caption{Loss Curve for TVSum-FM}
\label{fig:tvsum-fm}
\end{figure}

\begin{figure}[H]
    \begin{center}
        \includegraphics[width=75mm,scale=0.5]{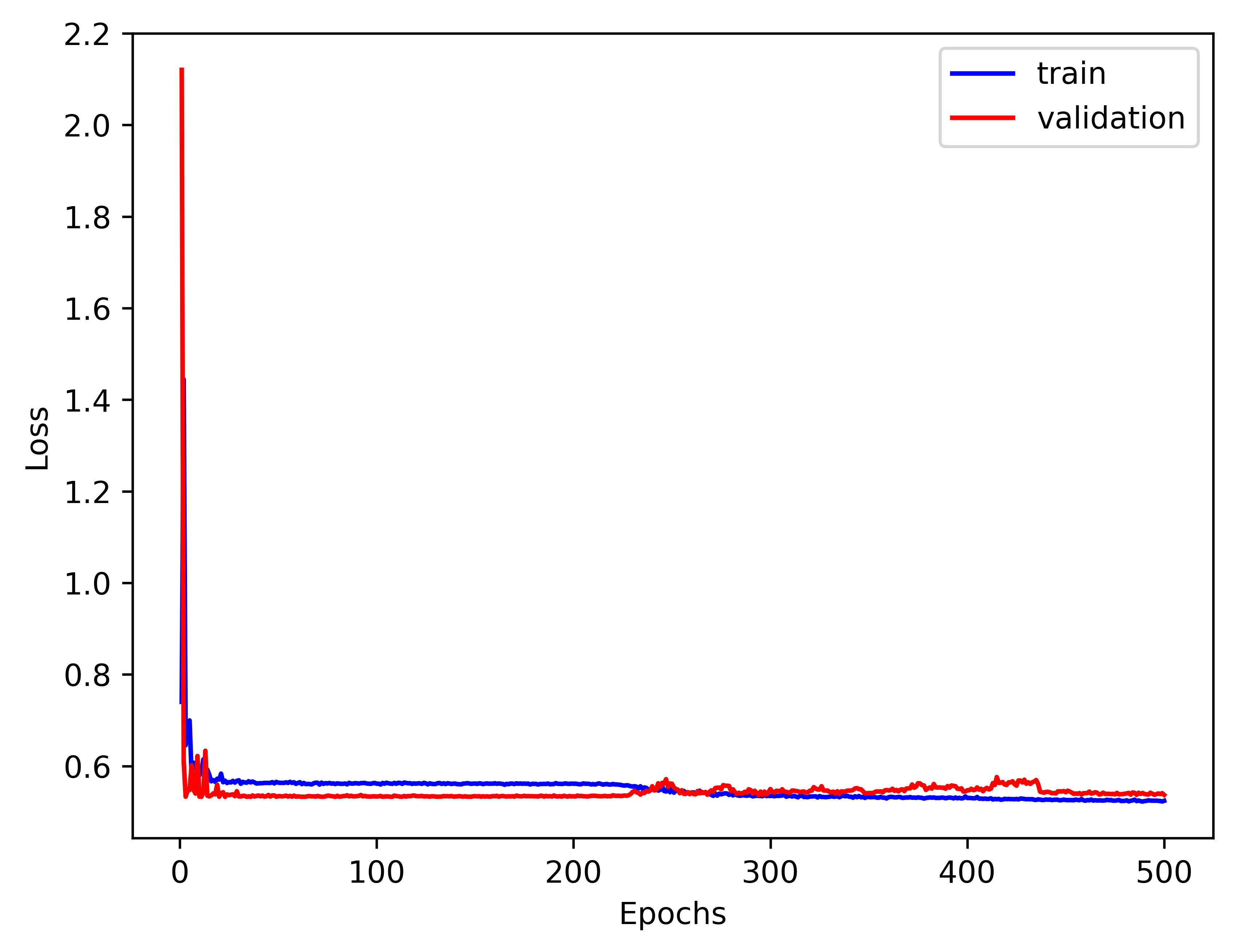}\\
    \end{center}
    \caption{Loss Curve for TVSum-GA}
\label{fig:tvsum-ga}
\end{figure}

\begin{figure}[H]
    \begin{center}
        \includegraphics[width=75mm,scale=0.5]{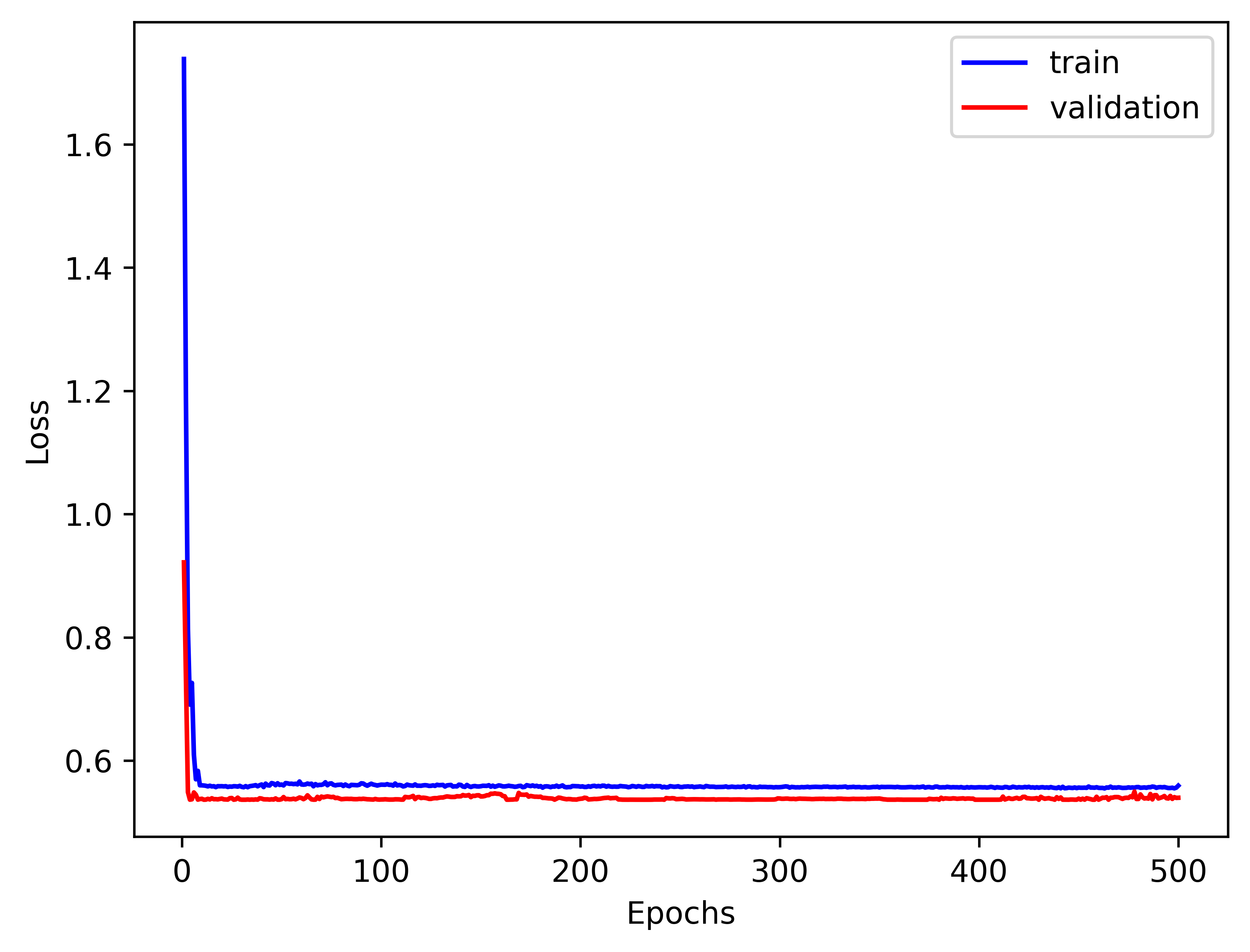}
    \end{center}
    \caption{Loss Curve for TVSum-MS}
\label{fig:tvsum-ms}
\end{figure}


\begin{figure}[H]
    \begin{center}
        \includegraphics[width=75mm,scale=0.5]{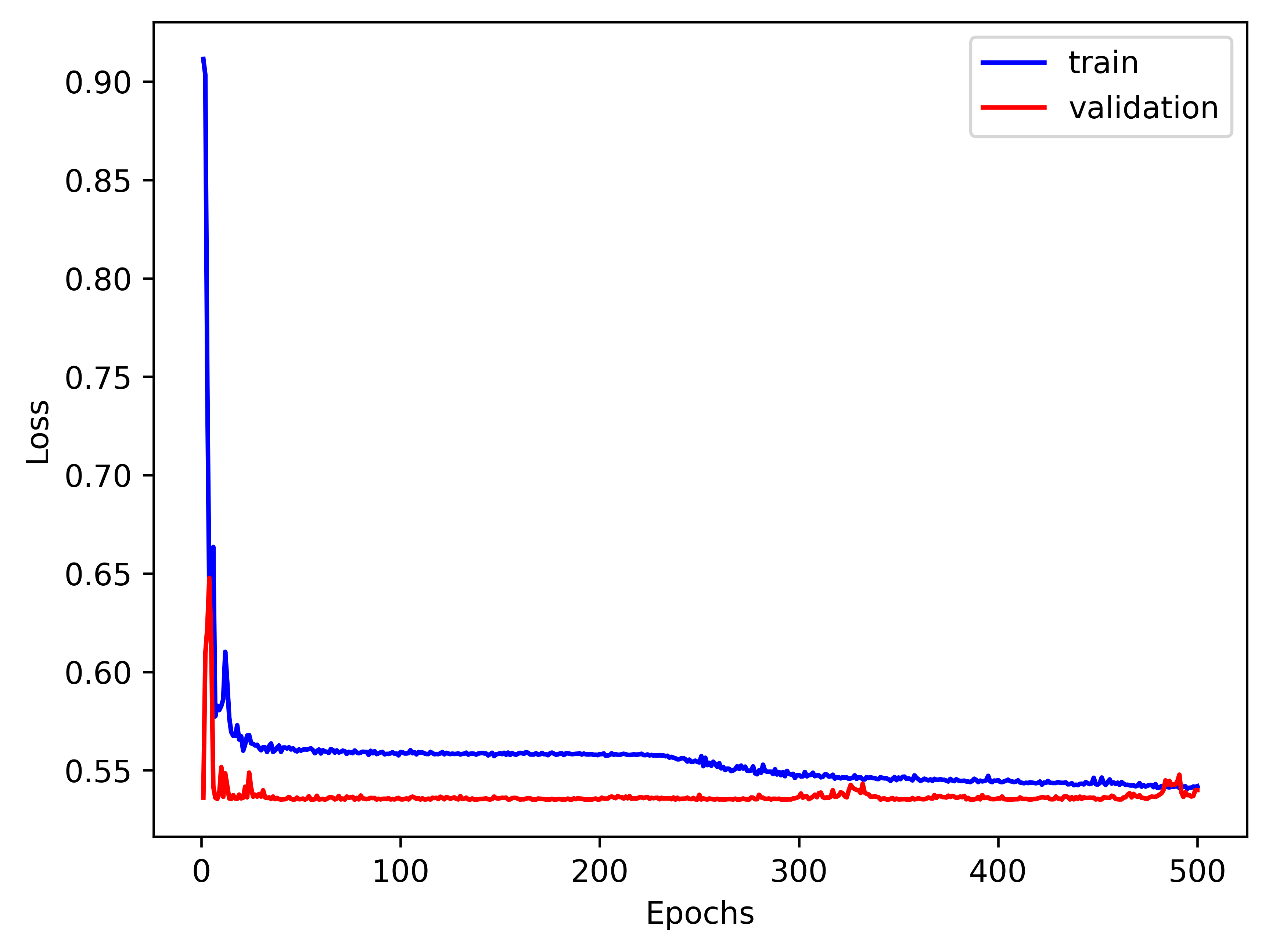}
    \end{center}
    \caption{Loss Curve for TVSum-PK}
\label{fig:tvsum-pk}
\end{figure}

\begin{figure}[H]
    \begin{center}
        \includegraphics[width=75mm,scale=0.5]{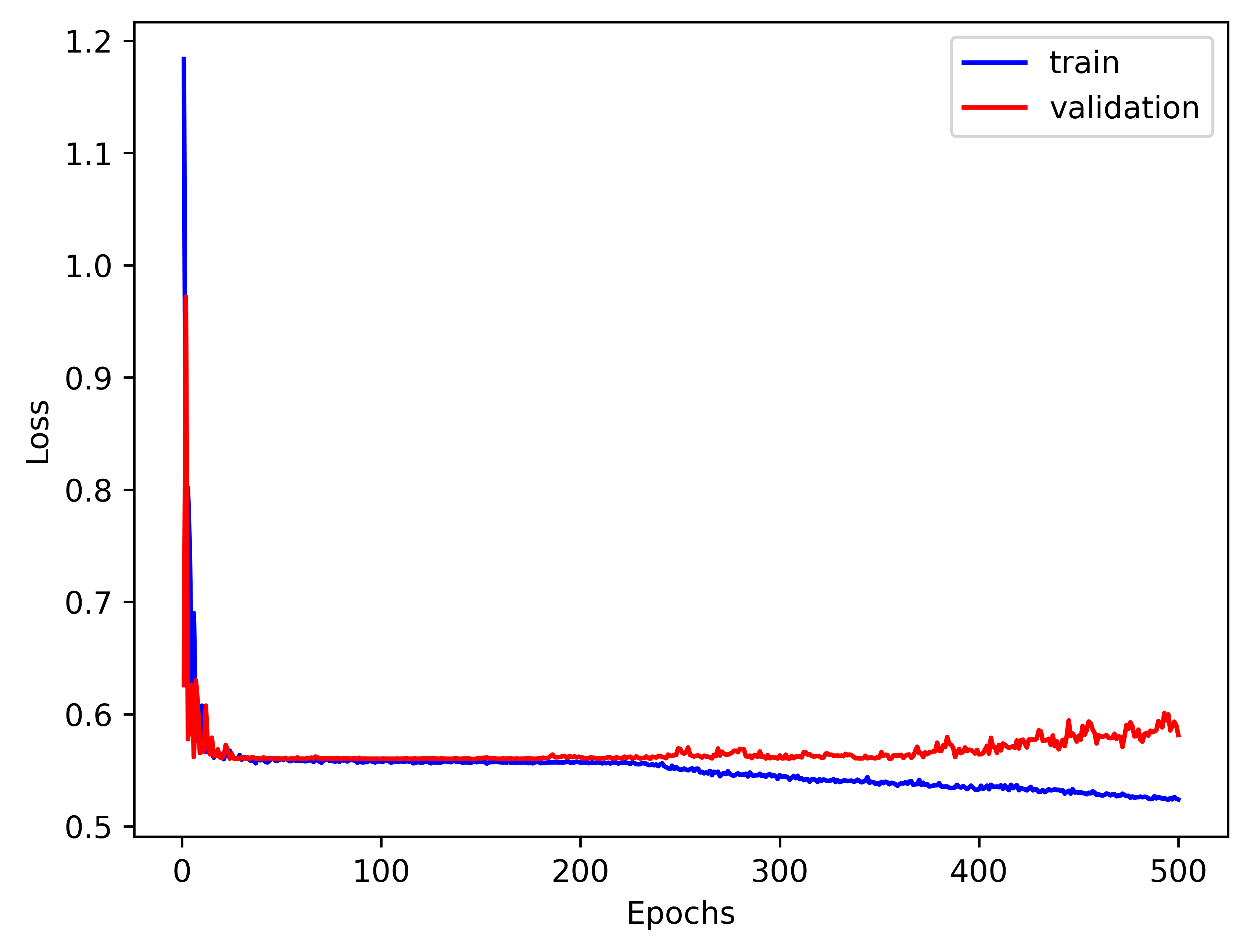}
    \end{center}
    \caption{Loss Curve for TVSum-PR}
\label{fig:tvsum-pr}
\end{figure}

\begin{figure}[H]
    \begin{center}
        \includegraphics[width=75mm,scale=0.5]{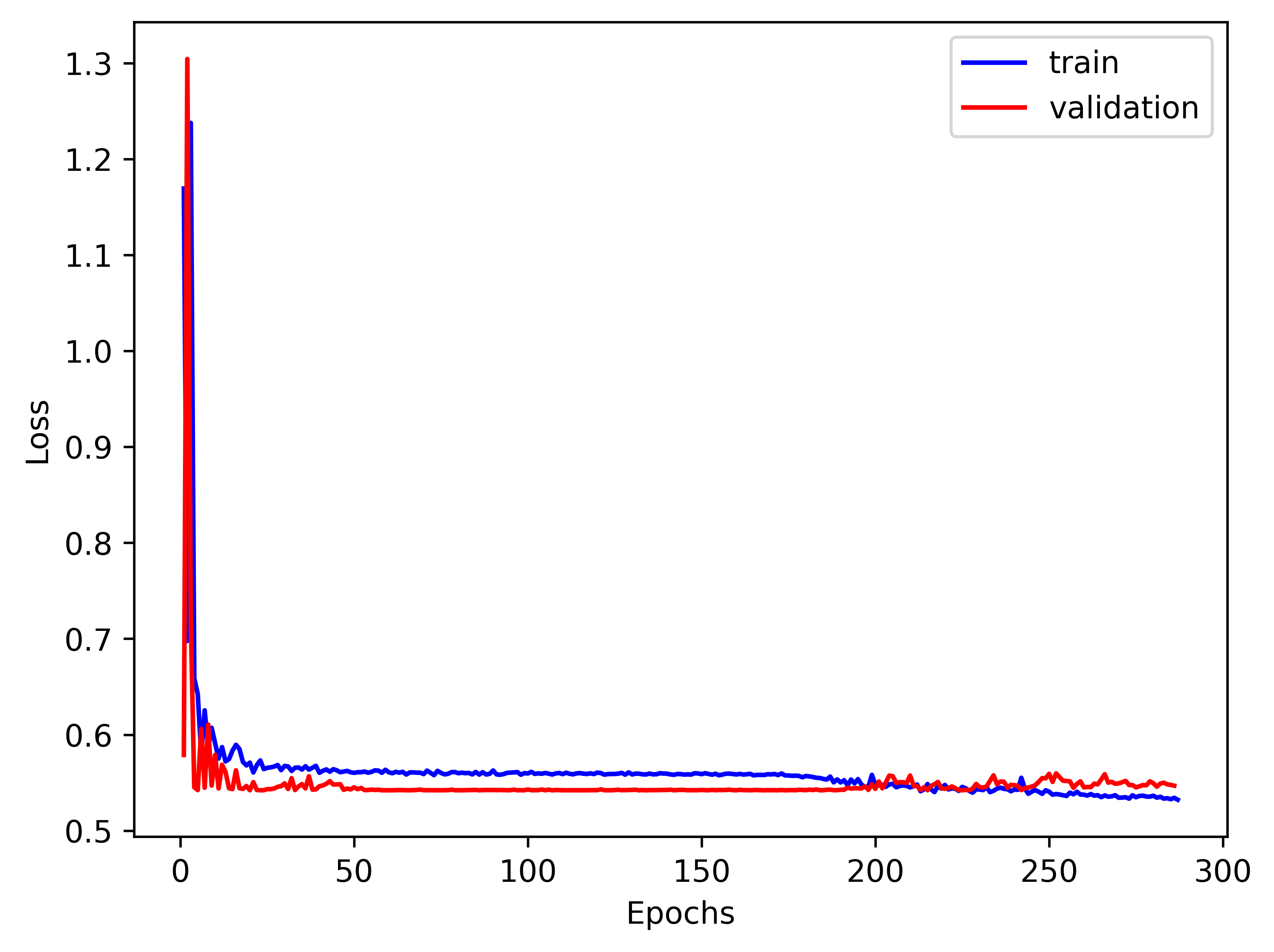}
    \end{center}
    \caption{Loss Curve for TVSum-VT}
\label{fig:tvsum-vt}
\end{figure}

\begin{figure}[H]
    \begin{center}
        \includegraphics[width=75mm,scale=0.5]{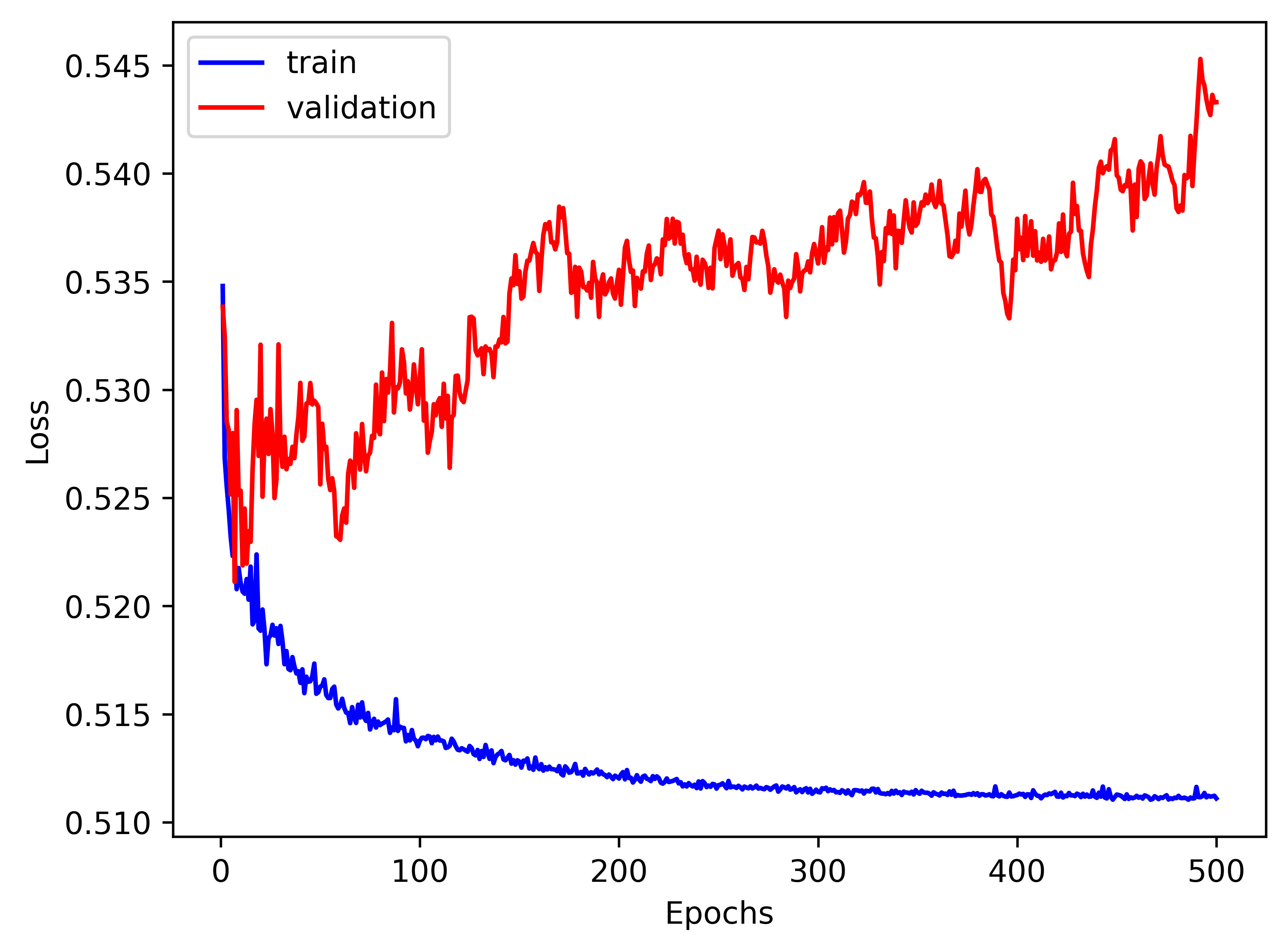}
    \end{center}
    \caption{Loss Curve for TVSum-VU}
\label{fig:tvsum-vu}
\end{figure}

\begin{figure}[H]
    \begin{center}
        \includegraphics[width=75mm,scale=0.5]{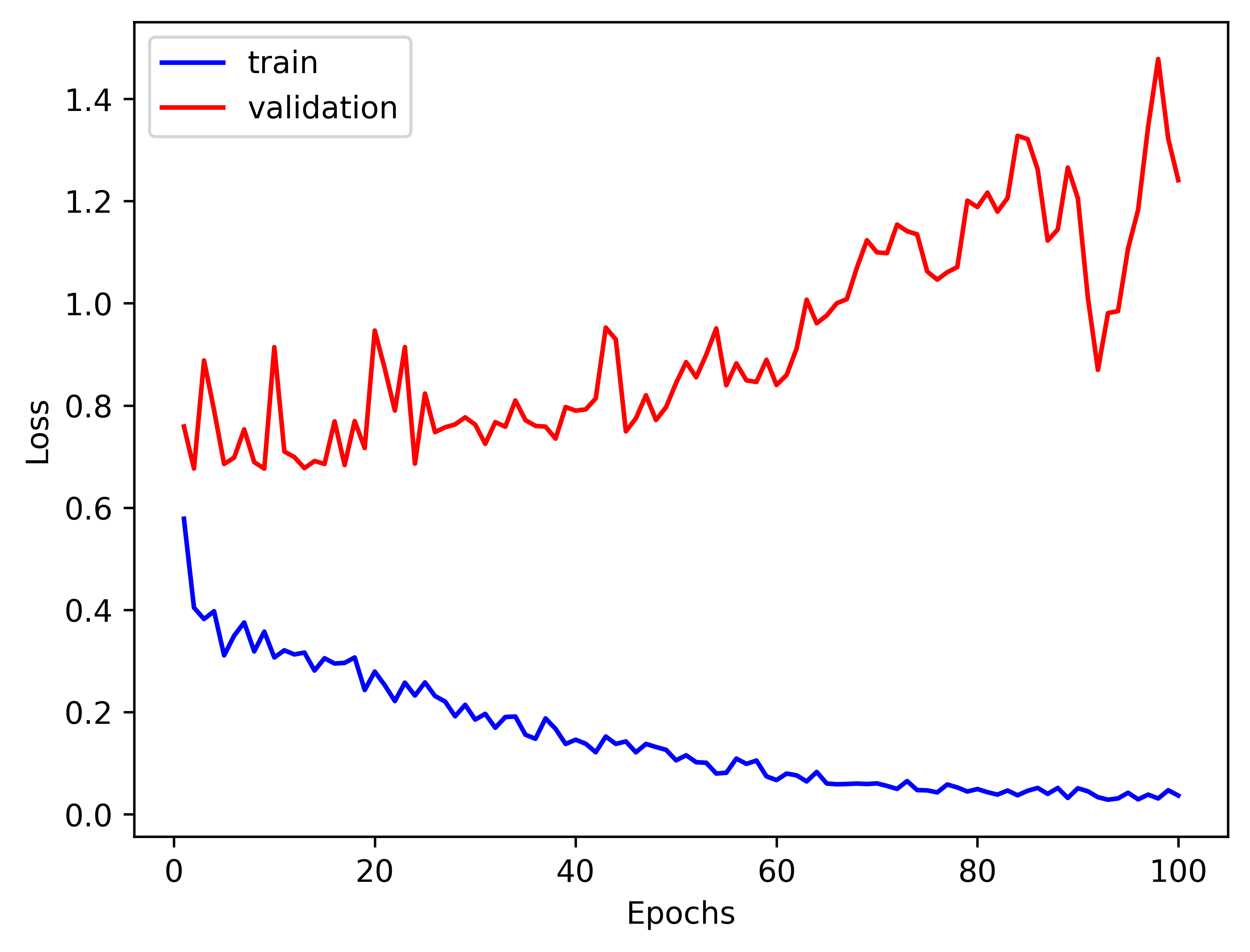}
    \end{center}
    \caption{Loss Curve for YT-Dog}
\label{fig:yt-dog}
\end{figure}

\begin{figure}[H]
    \begin{center}
        \includegraphics[width=75mm,scale=0.5]{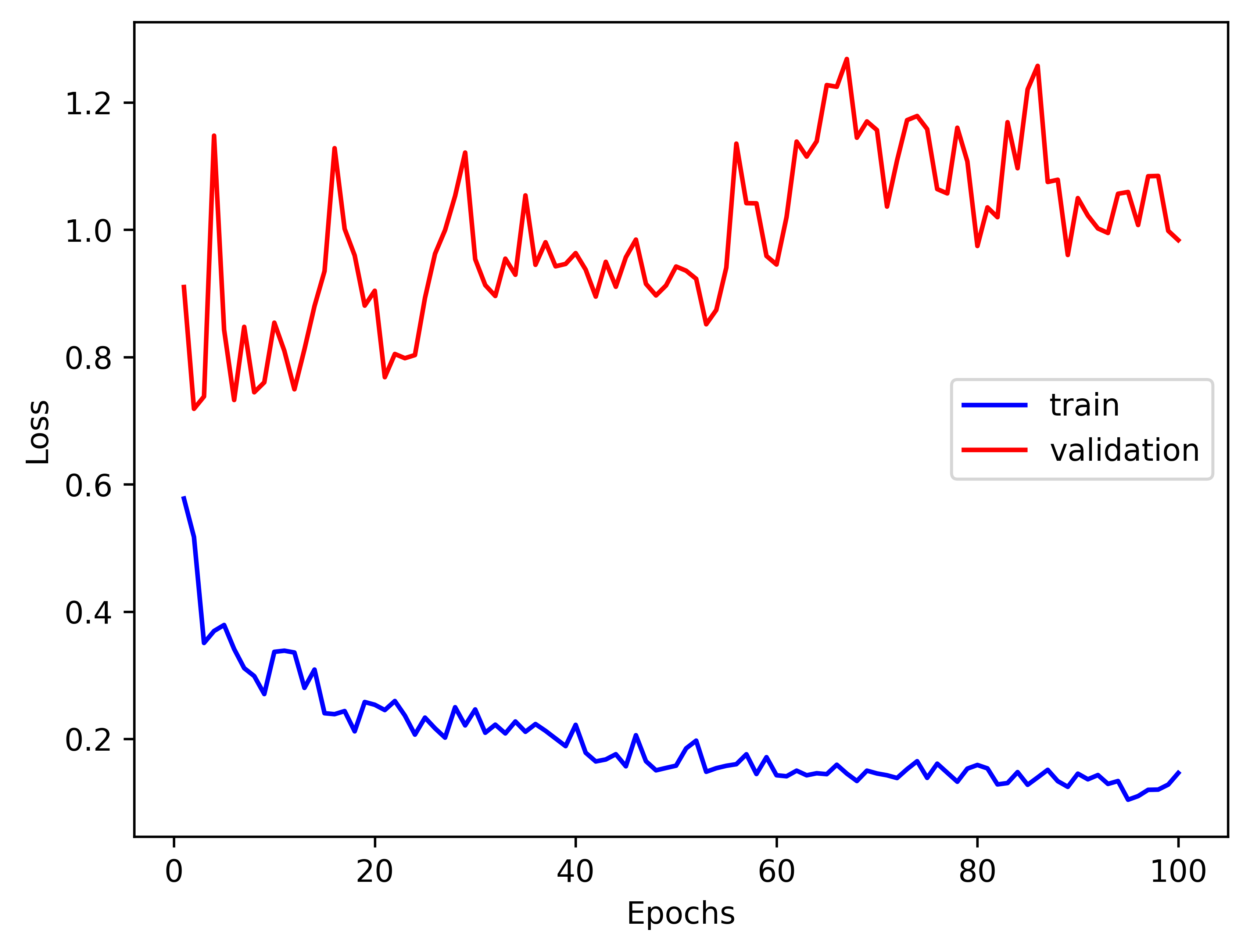}
    \end{center}
    \caption{Loss Curve for YT-Gym}
\label{fig:yt-gym}
\end{figure}

\begin{figure}[H]
    \begin{center}
        \includegraphics[width=75mm,scale=0.5]{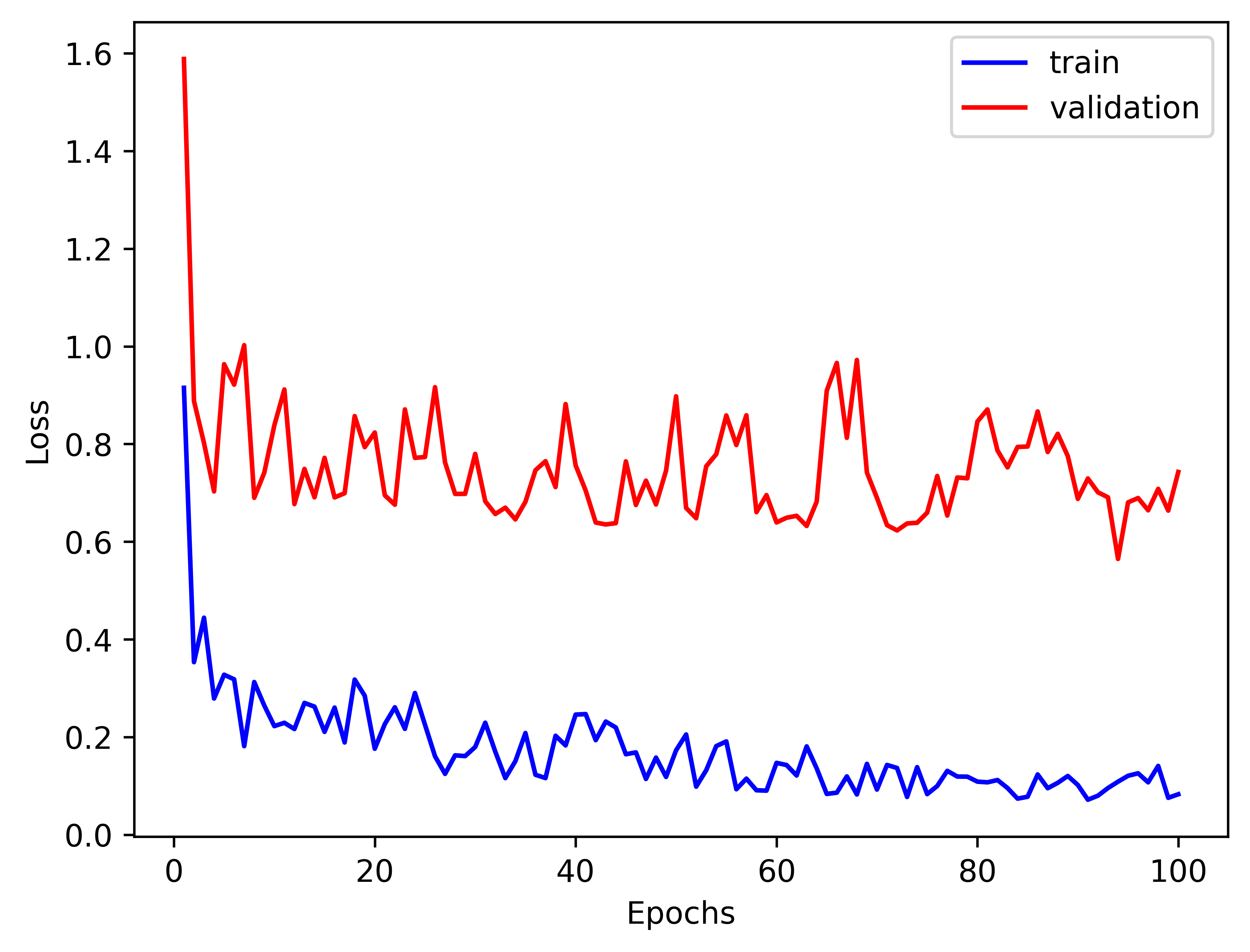}
    \end{center}
    \caption{Loss Curve for YT-Par}
\label{fig:yt-par}
\end{figure}

\begin{figure}[H]
    \begin{center}
        \includegraphics[width=75mm,scale=0.5]{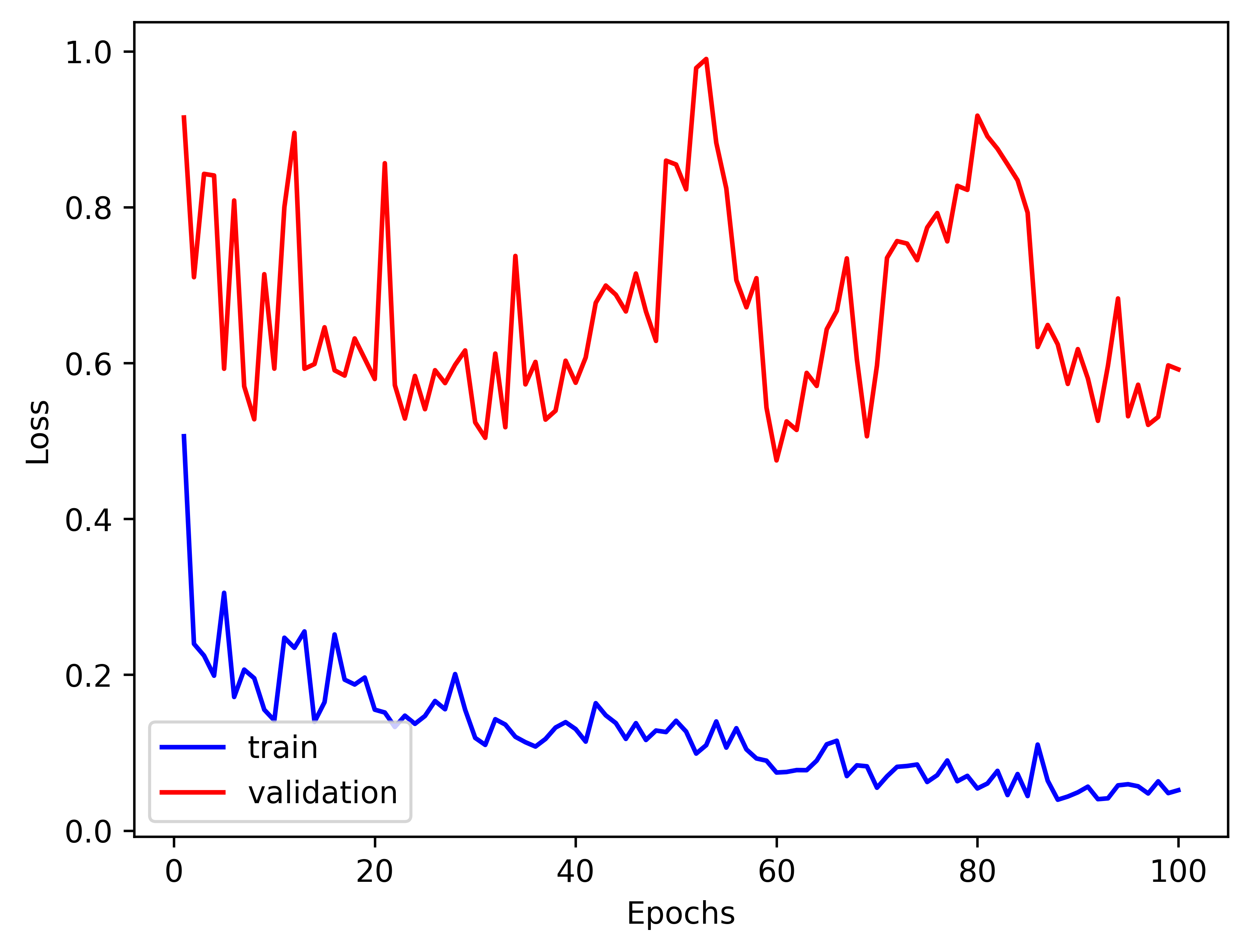}
    \end{center}
    \caption{Loss Curve for YT-Ska}
\label{fig:yt-ska}
\end{figure}

\begin{figure}[H]
    \begin{center}
        \includegraphics[width=75mm,scale=0.5]{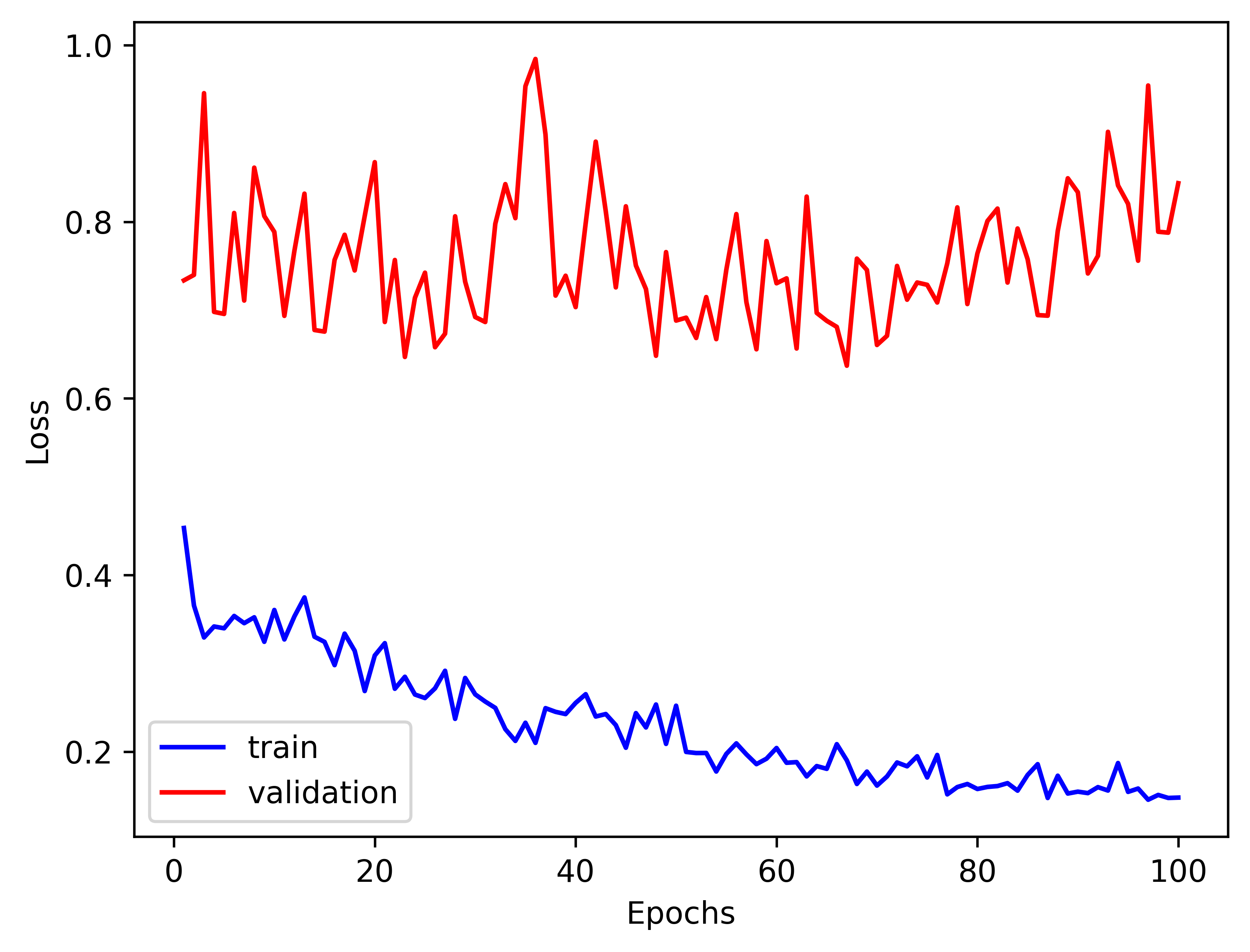}
    \end{center}
    \caption{Loss Curve for YT-Ski}
\label{fig:yt-ski}
\end{figure}

\begin{figure}[H]
    \begin{center}
        \includegraphics[width=75mm,scale=0.5]{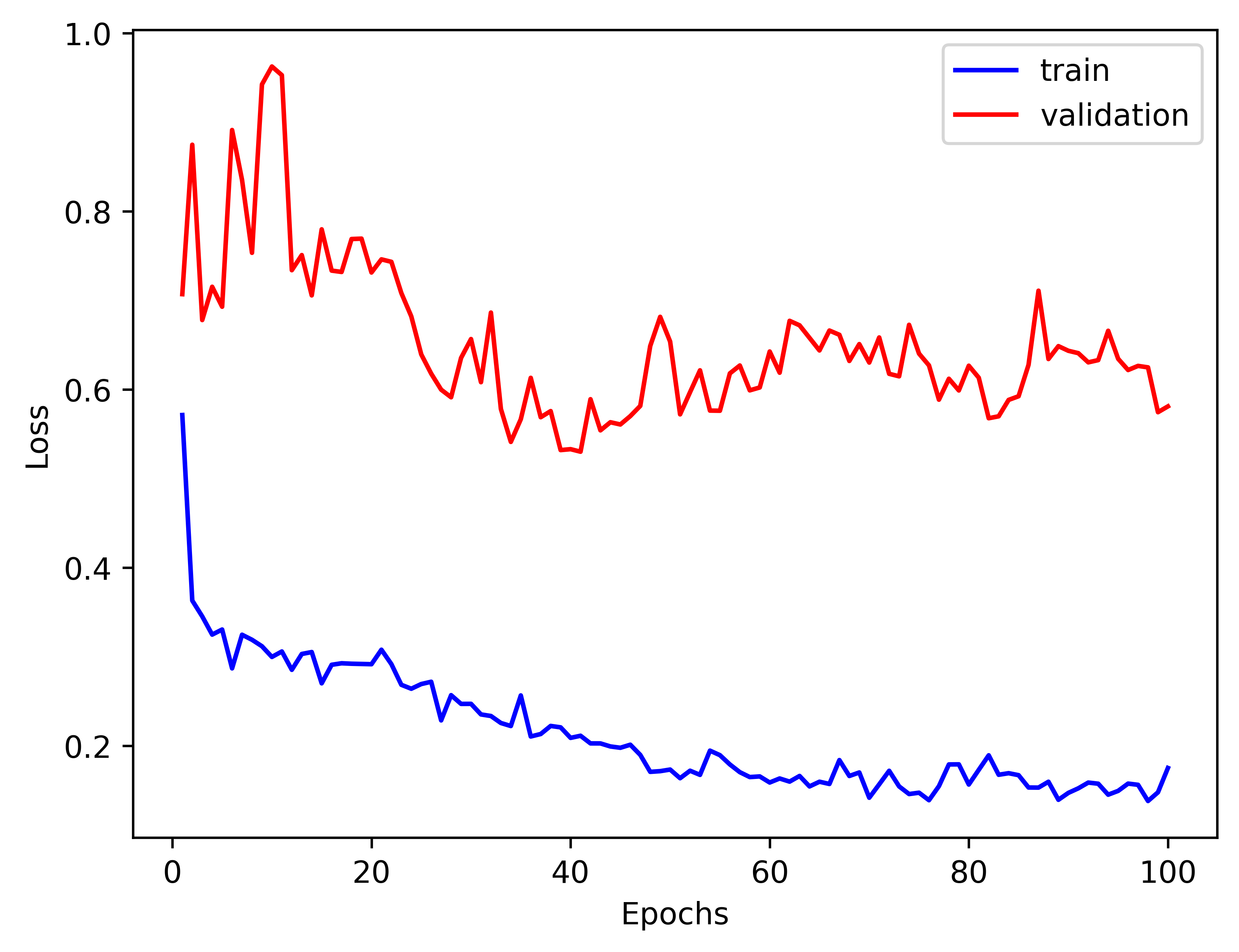}
    \end{center}
    \caption{Loss Curve for YT-Sur}
\label{fig:yt-sur}
\end{figure}

\end{document}